\documentclass[11pt]{amsart}

\usepackage[
            CJKbookmarks=true,
            bookmarksnumbered=true,
            bookmarksopen=true,
            colorlinks=true,
            citecolor=red,
            linkcolor=blue,
            anchorcolor=red,
            urlcolor=blue,
            pdfauthor={Marius Junge and Kiryung Lee},
            pdfstartview=FitH,
            ]{hyperref}

\usepackage{amsmath,amsxtra,amssymb,amsthm,amsfonts}
\usepackage[usenames]{color}
\usepackage{umoline}
\usepackage[all]{xy}
\usepackage{bm}
\usepackage{bbm}
\usepackage{tikz-cd}
\usepackage{epstopdf}
\usepackage{multirow}
\usepackage{graphicx}
\usepackage[caption=false]{subfig}

\newtheorem{lemma}{Lemma}[section]

\newtheorem{prop}[lemma]{Proposition}
\newtheorem{theorem}[lemma]{Theorem}
\newtheorem{cor}[lemma]{Corollary}

\newtheorem{rem}[lemma]{Remark}

\newcommand{\re}{\begin{rem}\rm}
  \newcommand{\mar}{\end{rem}}
\newtheorem{exam}[lemma]{Example}

\newtheorem{defi}[lemma]{Definition}

\newcommand{\kla}{\left ( }
\newcommand{\mer}{\right ) }

\newcommand{\fo}{\begin{eqnarray*}}
\newcommand{\mel}{\end{eqnarray*}}

\newcommand{\kl}{\pl \le \pl}
\newcommand{\gl}{\pl \ge \pl}

\newcommand{\lel}{\pl = \pl}

\newcommand{\ez}{{\mathbb E}}
\newcommand{\nz}{{\rm  I\! N}}

\newcommand{\rz}{{\mathbb R}}
\newcommand{\zz}{{\mathbb Z}}

\newcommand{\cz}{{\mathbb C}}
\newcommand{\kz}{{\rm  I\! K}}

\newcommand{\ten}{\otimes}

\newcommand{\pl}{\hspace{.1cm}}

\newcommand{\Om}{\Omega}

\newcommand{\al}{\alpha}
\newcommand{\si}{\sigma}

\newcommand{\La}{\Lambda}
\newcommand{\la}{\lambda}
\newcommand{\eps}{\varepsilon}

\newcommand{\E}{{\mathcal E}}

\DeclareMathOperator{\absc}{absconv}
\DeclareMathOperator{\Sh}{Sh}
\DeclareMathOperator{\Orb}{Orb}
\DeclareMathOperator*{\Motimes}{\text{\raisebox{0.25ex}{\scalebox{0.8}{$\bigotimes$}}}}

\newcommand{\qd}{\end{proof}\vspace{0.5ex}}

\newcommand{\xspace}{\hbox{\kern-2.5pt}}

\begin{document}

\title[Generalized notions of sparsity and RIP. Part I: A unified framework]{Generalized notions of sparsity and restricted isometry property. Part I: A unified framework}

\author{Marius Junge and Kiryung Lee}
\maketitle

\begin{abstract}
  The restricted isometry property (RIP) is an integral tool in the analysis of various inverse problems with sparsity models. Motivated by the applications of compressed sensing and dimensionality reduction of low-rank tensors, we propose generalized notions of sparsity and provide a unified framework for the corresponding RIP, in particular when combined with isotropic group actions. Our results extend an approach by Rudelson and Vershynin to a much broader context including commutative and noncommutative function spaces. Moreover, our Banach space notion of sparsity applies to affine group actions. The generalized approach in particular applies to high order tensor products.

\end{abstract}

\section{Introduction}

The {\em restricted isometry property} (RIP) has been used as a universal tool in the analysis of many modern inverse problems with sparsity prior models.  Indeed, the RIP implies that certain linear maps act as near isometries when restricted to ``nice'' (or sparse) vectors. Motivated from emerging big data applications such as compressed sensing or dimensionality reduction of massively sized data with a low-rank tensor structure, we provide a unified framework for the RIP allowing a generalized notion of sparsity and extend the existing theory to a much broader context.

Let us recall that in compressed sensing the RIP played a crucial role in providing guarantees for the recovery of sparse vectors from a small number of observations.  Moreover, these guarantees were achieved by practical polynomial time algorithms (e.g., \cite{candes2005decoding,rudelson2008sparse}).
In machine learning, the RIP enabled a fast and guaranteed dimensionality reduction of data with a sparsity structure.
The notion of sparsity has been shown for various sparsity models and in many cases the RIP turns out to be nearly optimal in terms of scaling of parameters for several classes of random linear operators.  For example, a linear map with random subgaussian entries satisfies a near optimal RIP for the canonical sparsity model \cite{candes2005decoding,baraniuk2008simple,krahmer2014suprema}, low-rank matrix model \cite{recht2010guaranteed,candes2011tight}, and low-rank tensor model \cite{rauhut2016low}. Baraniuk et al. \cite{baraniuk2008simple} provided an alternative elementary derivation that combines exponential concentration of a subgaussian quadratic form and standard geometric argument with union bounds.

Linear operators with special structures such as subsampled Fourier transform arise in practical applications.  These structures are naturally given by the physics of applications (e.g., Fourier imaging) and subsampled versions of these structured linear operators can be implemented within existing physical systems.  Furthermore, structured linear operators also enable scalable implementation at low computational cost, which is highly desirable for dimensionality reduction.  It has been shown that a partial Fourier operator satisfies a near optimal RIP for the canonical sparsity model in the context of compressed sensing \cite{candes2006near,rudelson2008sparse,rauhut2010compressive}.  For another example, quantum tomography, the linear operator for randomly subsampled Pauli measurements was shown to satisfy a near optimal RIP for a low-rank matrix model \cite{liu2011universal}.

There are applications whose setup doesn't fit in the existing theory because the classical sparsity model does not hold and/or assumptions on the linear operator are not satisfied. Motivated by such applications, in this paper, we extend the notion of sparsity and RIP for structured linear operator in several ways described below.

\subsection{Generalized notion of sparsity}
\label{sec:gen_notion}
\begin{figure}
  \centering
  \subfloat{\includegraphics[width=0.25\textwidth]{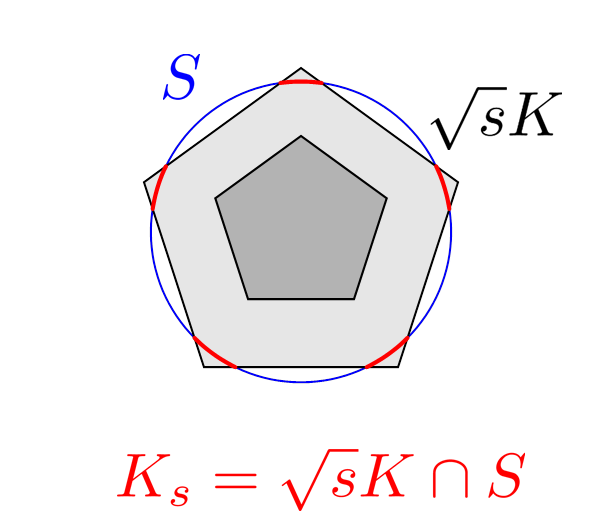}}
  \hspace{0.05\textwidth}
  \subfloat{\includegraphics[width=0.25\textwidth]{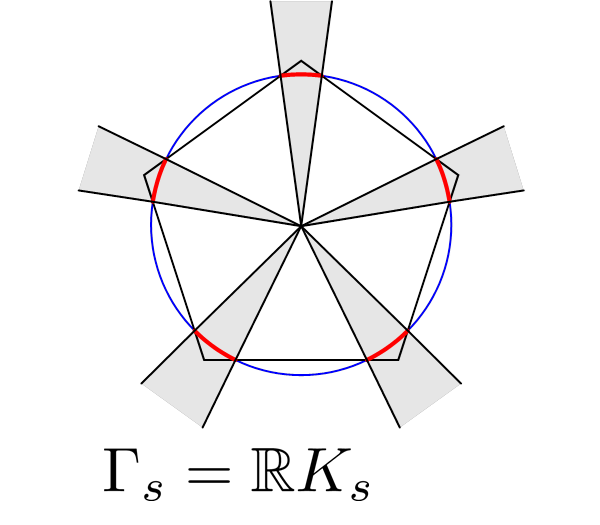}}
  \caption{Visualization of an abstract sparsity model using a convex set $K$ and the unit sphere $S$ in a Hilbert space $H$. Left: Set of $s$-sparse unit-norm vectors $K_s$ (red). Right: Set of $s$-sparse vectors $\Gamma_s$ (gray-shaded).}
  \label{fig:model}%
\end{figure}

First, we generalize the notion of sparsity.  Let $H$ be a Hilbert space and $K \subset H$ be a centered convex body.  We will consider the Banach space $(X,\|\cdot\|_X)$ obtained by completing the linear span of $K$ with the norm $\|\cdot\|_X$ given as the Minkowski functional $p_K(\cdot):X\to \mathbb{R}$ defined by $p_K(x) := \inf\{\la>0 \pl|\pl x\in \la K\}$.
\begin{defi}
We say that a vector $x\in H$ is $(K,s)$-sparse if
\[
\|x\|_X \leq \sqrt{s} \|x\|_H \pl ,
\]
where $X$ is the Banach space with unit ball $K$.
\end{defi}
The set of $(K,s)$-sparse unit-norm vector in $H$, denoted by $K_s$, is geometrically given as the intersection of $\sqrt{s} K$ and the unit sphere $S = \{ x \pl|\pl \|x\|_H = 1\}$.  Then the set of $(K,s)$-sparse vectors, denoted by $\Gamma_s$, is the star-shaped nonconvex cone given by $\mathbb{R} K_s$ (or $\mathbb{C} K_s$ if the scalar field is complex).  These two sets are visualized in Figure~\ref{fig:model}.  For example, if $H = \ell_2^N$ and $X = \ell_1^N$, then $\Gamma_s$ corresponds to the set of approximately $s$-sparse vectors with respect to the canonical basis.  The authors of this paper showed that existing near optimal RIP results extend from the exact canonical sparsity model to this {\em approximately} sparse model \cite{junge2015rip}.  This generalized notion of sparsity covers a wider class of models beyond the classical atomic model.  For example, in a companion Part II paper \cite[Section~4]{junge2016ripII}, we demonstrate a case where a sparse vector is not represented as a finite linear combination of atoms.  It also allows a machinery that optimizes sample complexity for the RIP of a given atomic sparsity model by choosing an appropriate Banach space (see \cite[Section~2]{junge2016ripII}).  In a special case, where the sparsity level $s$ is 1, our theory covers an arbitrary set.\footnote{Note that taking the convex hull of a given set does not increase the number of measurements for RIP.  Therefore, the convex set $K$ can be considered as the convex hull of a given set of interest in this case.}

\subsection{Vector-valued measurements}
\label{sec:vval}
Second, we consider vector-valued measurements which generalize the conventional scalar valued measurements.  This situation arises in several practical applications.  For example, in medical imaging and multi-dimensional signal acquisition, measurements are taken by sampling transform of the input not individually but in blocks.  The performance of $\ell_1$ norm minimization has been analyzed in this setup \cite{polak2015performance,bigot2016analysis} and it was shown that block sampling scheme, enforced by applications, adds a penalty to the number of measurements for the recovery.  This analysis extends the noiseless part of the analogous theory for the scalar valued measurements \cite{candes2011probabilistic}, which relies on a property called {\em local isometry}, which is a weaker version of the RIP.  For stable recovery from noisy measurements, one essentially needs the RIP of the measurement operator but block sampling setup does not fit to existing RIP results for structured linear operators.  In this paper, we will consider general vector-valued measurements in a Hilbert space and generalize the notion of incoherence and other properties accordingly.  This extension, in particular combined with a generalized sparsity model, requires the use of theory of factorization of a linear operator in Banach spaces \cite{pisier1986factorization}.

\subsection{Sparsity with enough symmetries and group structured RIP} $\atop$

We also generalize the theory of the RIP for partial Fourier measurement operators to more general group structured measurement operators, which will exploit the inherent structure in the Banach space that determines a sparsity model.
The canonical sparsity model is an examples of our general sparsity model in Section~\ref{sec:gen_notion}, where the convex set $K$ has a special structure called \emph{enough symmetries}.
Let $G$ be a group and $\si:G\to O_N$ be an affine representation that maps an element in $G$ to the orthogonal group in $\rz^N$.  An affine representation is isotropic if averaging the conjugate actions on any linear operator becomes a scalar multiple of the identity.
A convex set $K$ has enough symmetries if there exists an isotropic affine representation such that $\si(g) K = K$ for all $g \in G$.  Finite dimensional Banach spaces with enough symmetries have been studied extensively (see \cite{tomczak1989banach,defant1992tensor,pisier1986factorization}). Our original motivation for this problem comes from considering the low-rank tensor product in $\ell_2^n \otimes_\pi \ell_2^n \otimes_\pi \ell_2^n$. In fact a nice feature of spaces with enough symmetries comes from their stability under tensor products.

Under the enough symmetry of $K$, we consider a linear operator $v: X\to \ell_2^m$ given as the composition map $(v_j)_{1\leq j\leq m}$ given by sampling the (adjoint) orbit $\Orb(G)=\{\si(g)^*\eta \pl | \pl g\in G\}$ of $\eta \in X^*$, i.e. $v_j(x) = \langle \eta, \si(g_j) x\rangle$ for $x \in X$ and $j = 1,\dots,m$, where $(g_j)_{1\leq j\leq m} \subset G$. For certain class of groups, the group structured measurement operator $A = \frac{1}{\sqrt{m}} (v_j)_{1\leq j\leq m}$ has fast implementation. For example, if $\eta = [1,\dots,1]^\top \in \mathbb{R}^N$, then $A: \cz^N \to \cz^m$ reduces to a partial discrete Fourier transform. If the group actions consist of circular shifts in the canonical basis and in the Fourier basis, then $A$ corresponds to a partial quantum Fourier transform, which is a special case of the Gabor transform. We will demonstrate the RIP of this group structured measurement operators when the group elements are randomly selected.

Again, the group structured measurement operator is a natural extension of a partial Fourier operator. Unlike the other extension to subsampled bounded orthogonal system \cite{rauhut2010compressive}, the group structured measurement operator is tightly connected to a given sparsity model.


\subsection{Main results}

We illustrate our main results in the general setup with two concrete examples in the two theorems below.
These theorems provide the RIP of a random group structured measurement operator respectively for the corresponding stylized sparsity models. Both theorems assume that $H = \ell_2^N$ and the convex set $K$, which determines the set of $s$-sparse vectors $\Gamma_s$, has enough symmetry with an isotropic affine representation $\sigma: G \to O_N$ and $X$ is the Banach space induced from $K$ so that $K$ is the unit ball in $X$ as before. A set of random measurements are obtained by using random group actions. Specifically, we assume that $g_1,\dots,g_m$ are independent copies of a Haar-distributed random variable $g$ on $G$.

The first theorem demonstrates our main result in the case where $K$ is a polytope given as an absolute convex hull of finitely many vectors.
\begin{theorem}[Polytope]
\label{randrv_simplified}
Suppose that $X$ be an $N$-dimensional Banach space where its unit ball $K$ is an absolute convex hull of $M$ points. Let $K_s$ be defined from $K$ as before and $u:X\to\ell_2^d$ satisfy $\mathrm{tr}(u^*u) = N$. Then
\[
\sup_{x\in K_s} \Big| \frac{1}{m} \sum_{j=1}^m |u(\si(g_j)x)|_2^2 - \|x\|_2^2 \Big|
\kl \max(\delta,\delta^2)
\]
holds with high probability for $m = O(\delta^{-2} s \|u:X\to\ell_2^d\|^2 (1+\ln m)(1+\ln md)^2 (1+\ln M))$.
\end{theorem}

Theorem~\ref{randrv_simplified} generalizes the RIP result of a partial Fourier operator (e.g., \cite{rudelson2008sparse}) in the three ways discussed above.  The operator norm of $v$ in Theorem~\ref{randrv_simplified} generalizes the notion of incoherence in existing theory.  Most interestingly, combined with a clever net argument, Theorem~\ref{randrv_simplified} enables the RIP of a random group structured measurement operator for low-rank tensors (See Section~\ref{sec:tensor}).

The second theorem deals with the sparsity model with respect to a ``nice'' Banach space whose norm dual has type $p$ \cite{pisier1999volume}.
(
Details are explained in Section~\ref{subsec:dualtypep}.) Here for simplicity we only demonstrate an example where $p=2$.
\begin{theorem}[Dual of type 2]
\label{point_simplified}
Suppose that $X$ is an $N$-dimensional Banach space such that its norm dual $X^*$ has type 2. Let $K_s$ be defined from the unit ball $K$ in $X$ as before and $\eta\in\cz^N$ satisfy $\|\eta\|_2=\sqrt{N}$. Then
\[
\sup_{x\in K_s} \Big|\frac{1}{m}\sum_{j=1}^m |\langle \eta, \si(g_j)x\rangle|^2 - \|x\|_2^2\Big| < \max(\delta,\delta^2) \pl
\]
holds with high probability for $m = O(\delta^{-2} s T_2(X^*)^6 \|\eta\|_{X^*} (1+\ln m)^3)$, where $T_2(X^*)$ denotes the type 2 constant of $X^*$.
\end{theorem}
Theorem~\ref{point_simplified} covers many known results on the RIP of structured random linear operator and should be considered as an umbrella result for this theory. Importantly Theorem~\ref{point_simplified} applies to noncommutative cases such as Schatten classes and the previous result for a partial Pauli operator applied to low-rank matrices \cite{liu2011universal} is a special example.

In fact, Theorems~\ref{randrv_simplified} and \ref{point_simplified} are just exemplar of the main result in full generality in Theorem~\ref{main:rv}. In the Part II paper \cite{junge2016ripII}, we also demonstrate that Theorem~\ref{main:rv} provides theory for the RIP for infinite dimensional sparsity models.

\subsection{Notation}

In this paper, the symbols $c,c_1,c_2,\dots$ and $C,C_1,C_2,\dots$ will be reserved for numerical constants, which might vary from line to line.
We will use notation for various Banach spaces and norms.
The operator norm will be denoted by $\|\cdot\|$.
We will use the shorthand notation $\|\cdot\|_p$ for the $\ell_p$-norm for $1\leq p\leq \infty$.
For $N \in \nz$, the unit ball in $\ell_p^N$ will be denoted by $B_p^N$.
For set $\mathcal{I} \subset \zz$, let $(\bm{e}_j)_{j\in\mathcal{I}}$ denote the canonical basis for $\cz^{|\mathcal{I}|}$. The index set $\mathcal{I}$ should be clear from the context.
The identity operator will be denoted by $\mathrm{Id}$.
For set $\mathcal{S}$ of linear operators, the commutant, denoted by $\mathcal{S}'$, refers to the set of linear operators those commute with all elements in $\mathcal{S}$, i.e.
\[
\mathcal{S}' = \{ T \pl|\pl TS = ST, \pl \forall S \in \mathcal{S}\}.
\]

\subsection{Organization}

The rest of this paper is organized as follows.
The main theorems are proved in Section~\ref{sec:rudver}.
We discuss the complexity of the convex set $K$ for various sparsity models in Section~\ref{sec:complexity}.
After a brief review of affine group representations and enough symmetries in Section~\ref{sec:symmetry}, by collecting the results from the previous sections, we illustrate the implication of the main results for prototype sparsity models in Section~\ref{sec:grmeas}.
Finally, we conclude the paper with the application of the main results for a low-rank tensor model in Section~\ref{sec:tensor}.

\section{Rudelson-Vershynin method}
\label{sec:rudver}

In this section, we derive a unified framework that identifies a sufficient number of measurements for the RIP of structured random operators in the general setup introduced in Sections~\ref{sec:gen_notion} and \ref{sec:vval}. We will start with the statement of the property in the general setup, followed by the proof.

\subsection{RIP in the general setup}
\label{sec:setup}
Let $H$ be a Hilbert spaces and $K \subset H$ be a centrally symmetric convex set and $X$ be the Banach space with unit ball $K$ as before. Let $\Gamma_s = \{ x \in X \pl|\pl \|x\|_X \leq \sqrt{s} \|x\|_H \}$ denote the set of $(K,s)$-sparse vectors and $K_s$ be the intersection of $\Gamma_s$ and the unit sphere in $H$. Let $v_1,\dots,v_m$ be independent random linear operator from $X$ to $\ell_2^d$. For notational simplicity, we let $v:X\to \ell_\infty^m(\ell_2^d)$ denote the composite map $(v_j)_{1\leq j\leq m}$ defined by $(v_j)_{1\leq j\leq m}(x) = (v_j(x))_{1\leq j \leq m}$ for $x \in X$. Then the measurement operator $A:X\to \ell_\infty^m(\ell_2^d)$, defined by $A = \frac{1}{\sqrt{m}} (v_j)_{1\leq j\leq m}$, generates a set of $m$ vector valued linear measurements in $\ell_2^d$.

Our results are stated for a class of incoherent random measurement operators. We adopt the arguments by Candes and Plan \cite{candes2011probabilistic} to describe these measurement operators. In the special case of $X = \ell_1^N$ and $d=1$, Candes and Plan considered a class of linear operators given by measurement maps satisfying the following two key properties. i) Isotropy: $\ez v_j^* v_j = I_N$ for all $j = 1,\dots,m$, where $I_N$ is the identity matrix of size $N$; ii) Incoherence: $\|v_j\|_\infty$ is upper bounded by a numerical constant $\mu$ (deterministically or with high probability). In our setup, the isotropy extends to
\begin{equation}
\label{isotropy}
\ez v_j^* v_j = \mathrm{Id} \pl .
\end{equation}
But sometimes we will also consider the case where
\begin{equation}
\label{common}
\ez v_j^* v_j = \Phi, \quad \forall j=1,\dots,m,
\end{equation}
hold with $\Phi: H\to H$ satisfying
\begin{equation}
\label{normalization}
\|\Phi\| \leq 1 \pl .
\end{equation}
Obviously, the isotropy is a sufficient condition for the relaxed properties in \eqref{common} and \eqref{normalization}. For the incoherence, we generalize it using an 1-homogeneous function $\al_d: B(X,\ell_2^d)\to [0,\infty)$ that maps a bounded map from $X$ to $\ell_2^d$ to a nonnegative number. A natural choice of $\al_d$ is the operator norm, which is consistent with the above example of $K = B_1^N$ and $d = 1$. The operator norm of $v_j$ in this case reduces to $\|v_j\|_{X^*} = \|v_j\|_\infty$. However, in certain scenarios, there exists a better choice of $\al_d$ than the operator norm that further reduces the sample complexity that identifies a sufficient number of measurements for the RIP. One such example is demonstrated for the windowed Fourier transform in the Part II paper \cite[Section~2]{junge2016ripII}.

Under the relaxed isotropy conditions in \eqref{common} and \eqref{normalization}, with a slight abuse of terminology, we say that $A$ satisfies the RIP on $\Gamma_s$ with constant $\delta$ if
\begin{equation}
\label{rip}
\sup_{x\in K_s} \Big| \|Ax\|_{\ell_2^m(\ell_2^d)}^2 - \langle x, \Phi x\rangle \Big|
\kl \max(\delta,\delta^2) \pl.
\end{equation}
In the special case where the isotropy ($\Phi = \mathrm{Id}$) is satisfied, the deviation inequality in \eqref{rip} reduces to the conventional RIP. Note that $\Phi$ is a nonnegative operator by construction. If $\Phi$ is a positive operator, then $\langle x, \Phi x\rangle$ is a weighted norm of $x$ and \eqref{rip} preserves this weighted norm through $w$ with a small perturbation proportional to $\|x\|_{H}^2$.

Our main result is a far reaching generalization of the RIP of a partial Fourier operator by Rudelson and Vershynin \cite{rudelson2008sparse}. We adapt their derivation that consists of the following two steps: The first step is to show that the expectation of the restricted isometry constant is upper bounded by the $\gamma_2$ functional \cite{talagrand1996majorizing} of the restriction set, then by an integral of the metric entropy number by Dudley's theorem \cite{ledoux2013probability}. Later in this section, we show that the first step extends to the general setup with the upper bound given by
\begin{equation}
\label{dudley}
\int_0^\infty \sqrt{\ln N(v(K),B_{\ell_\infty^m(\ell_2^d)})} \pl d\epsilon
\lesssim \sum_{l=0}^\infty \frac{e_l(v)}{\sqrt{l}} =: \E_{2,1}(v) \pl ,
\end{equation}
where $e_l(v)$ denotes the dyadic entropy number of $v:X\to\ell_\infty^m(\ell_2^d)$ \cite{carl1990entropy}.
The second step is where our theory deviates significantly from the previous work \cite{rudelson2008sparse}. Rudelson and Vershynin used a variation of Maurey's empirical method \cite{carl1985inequalities} to get an upper bound on the integral in \eqref{dudley} for $K$ being the unit ball in $\ell_1^N$, which in turn provided a near optimal sample complexity up to a logarithmic factor.
Liu \cite{liu2011universal} later extended the result by Rudelson and Vershynin \cite{rudelson2008sparse} to the case of a partial Pauli operator applied to low-rank matrices via the dual entropy argument by Gu{\'e}don et al. \cite{guedon2008majorizing}.

Our result is further generalization of these results. In particular, our result provides flexibility that can address the vector-valued measurement case and optimize sample complexity over the choice of the 1-homogeneous function $\al_d$ on $B(X,\ell_2^d)$. In the general setup, we need to adopt other tools in Banach space theory to get an analogous upper bound.
For this purpose, we introduce a property of the convex set $K$, defined as follows: We say that $K$ is of {\em entropy type $(p,\al_d)$} if there exists a constant $M_{p,\al_d}(K)$ such that
\begin{equation}
\label{Mpal}
\E_{2,1}(v) \kl M_{p,\al_d}(K) m^{1/2-1/p} (1+\ln m)^{e(p)/2} \sup_{1\leq j\leq m} \al_d(v_j)
\end{equation}
holds for any composite map $v = (v_j)_{1\leq j\leq m}$, where the exponent function $e(\cdot)$ is defined by $e(p) = 1$ for $1 \leq p < 2$ and $e(p) = 3$ for $p = 2$ for technical reasons. In this paper, $M_{p,\al_d}(K)$ will denote the smallest constant that satisfies \eqref{Mpal}. Note that $\al_d$ generalizes the notion of incoherence and $M_{p,\al_d}$ represents the complexity of a given sparsity model, which is discussed in more details in Section~\ref{sec:complexity}.

Our main theorem below identifies a sufficient number of measurements for RIP of random linear operator in the general setup.

\begin{theorem}
\label{main:rv}
Let $H$, $K$, $X$, $\Gamma_s$ be defined as above. Suppose that $K$ is of entropy type $(p,\al_d)$.
Let $v_1,\dots,v_m$ be independent random maps from $X$ to $\ell_2^d$ satisfying \eqref{common} and \eqref{normalization}.
Let $1 \leq p \leq 2$ and $0 < \zeta < 1$.
Then there exists a numerical constant $c$ such that $A = \frac{1}{\sqrt{m}} (v_j)_{1\leq j \leq m}$ satisfies the RIP on $\Gamma_s$ with constant $\delta$ with probability $1-\zeta$ provided
\begin{align}
\label{sample_comp1}
\frac{m^{1/p}}{(1+\ln m)^{e(p)/2}} &\gl c M_{p,\al_d}(K) \sqrt{s} \delta^{-1} \sup_{k \in \mathbb{N}} \Big(\ez \sup_{1 \leq j \leq m} \al_d(v_j)^{2k}\Big)^{1/2k}
\intertext{and}
\label{sample_comp2}
m &\gl c \delta^{-2} s \ln(\zeta^{-1}) \sup_{k \in \mathbb{N}} \Big(\ez \sup_{1 \leq j \leq m} \|v_j\|^{2k}\Big)^{1/k} \pl .
\end{align}
\end{theorem}

The moment terms in \eqref{sample_comp1} and \eqref{sample_comp2} are essentially probabilistic or deterministic upper bounds on $\sup_{1 \leq j \leq m} \al_d(v_j)$ and $\sup_{1 \leq j \leq m} \|v_j\|$, respectively.\footnote{Indeed, a tail bound implies moment bounds by the Markov inequality and the converse can be shown by direct calculation with Stirling's approximation of the gamma function. (e.g., see \cite[Chapter~7]{foucart2013mathematical}.)} These two terms extend the notion of incoherence of measurement functionals with respect to the given sparsity model. On the other hand, $M_{p,\al_d}(K)$ describes the complexity of sparsity model. In many of well-known examples, $M_{p,\al_d}(K)$ reduces to a logarithmic factor. However, if the convex set $K$ that determines the sparsity model has a bad geometry, there will be a penalty given by larger $M_{p,\al_d}(K)$. These incoherence and complexity parameters are controlled by a choice of the parameter $p$ and the 1-homogeneous function $\al_d$.

\begin{rem}{\rm
\label{rem:main:rv}
A natural choice for the parameters in Theorem~\ref{main:rv} is $p=2$ and $\al_d = \|\cdot\|$. Then the conditions in \eqref{sample_comp1} and \eqref{sample_comp2} reduces to
\[
m \gl c \delta^{-2} s \max\Big(M_{2,\|\cdot\|}^2 (1+\ln m)^3,\ln(\zeta^{-1})\Big) \sup_{k \in \mathbb{N}} \Big(\ez \sup_{1 \leq j \leq m} \|v_j\|^{2k}\Big)^{1/k} \pl .
\]
However, as shown in \cite[Section~2]{junge2016ripII}, there are cases where we can further reduce the number of measurements for the RIP in \eqref{sample_comp1} by optimizing over $K$, $p$, and $\al_d$.}
\end{rem}

\subsection{Proof of Theorem~\ref{main:rv}} $\atop$

Next we prove Theorem~\ref{main:rv}. In the course of the proof, we show that the Rudelson-Vershynin argument \cite{rudelson2008sparse} to derive a near optimal RIP of partial Fourier operators generalizes to a flexible method.
Let us start with recalling the relevant notation. Let $T: X \to Y$ be a linear map and $D$ be a subset of $X$. The dyadic entropy number \cite{carl1990entropy} is defined by
\[
e_l(T,D) := \inf\{\eps > 0 \pl | \pl N(T(D),\eps B_Y)\kl 2^{l-1}\} \pl .
\]
For $D=B_X$, we use the shorthand notation $e_l(T) = e_l(T,B_X)$.
The following equivalence between metric and dyadic entropy numbers is well known (see e.g., \cite{pisier1999volume}).
\begin{lemma}[\cite{pisier1999volume}]\label{ent}
$\int_0^{\infty} \sqrt{\ln N(T(D),\eps)} d\eps
\sim \sum_{l=1}^{\infty} \frac{e_l(T,D)}{\sqrt{l}}$.
\end{lemma}
Note that since $(e_l(T))_{l\in\mathbb{N}}$ is a nondecreasing sequence, $\sum_{l=1}^{\infty} \frac{e_l(T)}{\sqrt{l}}$ coincides with the norm of $(e_l(T))_{l\in\mathbb{N}}$ in the Lorentz sequence space $\ell(2,1)$ \cite{berg1976interpolation}. Therefore, we will use the shorthand notation $\E_{2,1}(T)$ to denote $\sum_{l=1}^{\infty} \frac{e_l(T)}{\sqrt{l}}$.

The following lemma provides a key estimate in proving Theorem~\ref{main:rv}.
\begin{lemma}\label{key}
Let $H$, $K$, $X$, and $K_s$ be defined as before. Let $D\subset K_s$. Let $v_1,\dots,v_m$ be linear maps from $X$ to $\ell_2^d$ and $v = (v_j)_{1\leq j\leq m}$ denote the composite map. Let $\xi_1,\dots,\xi_m$ be independent copies of $\xi \sim \mathcal{N}(0,1)$.
Then for all $k \in \mathbb{N}$
\[
\Big(\ez \sup_{x\in D} \Big|\sum_{j=1}^m \xi_j \|v_j(x)\|_2^2\Big|^k\Big)^{1/k}
\kl C \sqrt{s} \pl \Big(\sup_{x\in D} \sum_{j=1}^m \|v_j(x)\|_2^2\Big)^{1/2} \pl \Big(\E_{2,1}(v) + \sqrt{k} \|v\|\Big)\pl .
\]
\end{lemma}

\begin{proof}
Define $\beta(x):=(\|v_j(x)\|_2^2)_{1\leq j\leq m}$ for $x \in H$ and $\beta(D) := \{ \beta(x) \pl | \pl x\in D\}$. Let $\bm{\xi} = [\xi_1,\dots,\xi_m]^\top$.
Then $\langle \bm{\xi}, \beta(x)\rangle$ is a subgaussian process indexed by $x$.
By the tail bound result via generic chaining \cite[Theorem~3.2]{dirksen2015tail} and Dudley's inequality \cite{ledoux2013probability}, for all $k \in \mathbb{N}$ we have
\begin{equation}
\label{sjoerd}
\Big(\ez \sup_{x\in D} |\langle \bm{\xi}, \beta(x)\rangle|^k\Big)^{1/k}
\lesssim \int_0^{\infty}\sqrt{\ln N(\beta(D),\eps B_{\ell_2^m})} d\eps
+ \sup_{x\in D} \Big(\ez |\langle \bm{\xi}, \beta(x)\rangle|^k\Big)^{1/k} \pl .
\end{equation}

We first compute an upper bound on the first summand in the right-hand-side of \eqref{sjoerd}. Let
\begin{equation}
\label{eq:defR}
R := \Big(\sup_{x\in D} \sum_{j=1}^m \|v_j(x)\|_2^2\Big)^{1/2} \pl .
\end{equation}
Then we note that for $x,x'\in D$ we have
\begin{align*}
& \|\beta(x)-\beta(x')\|_2
\lel \Big(\sum_{j=1}^m |\|v_j(x)\|_2^2 - \|v_j(x')\|_2^2|^2\Big)^{1/2} \\
& \kl \Big(\sum_{j=1}^m |\langle v_j(x-x'),v_j(x)\rangle|^2\Big)^{1/2}
+ \Big(\sum_{j=1}^m |\langle v_j(x),v_j(x-x')\rangle|^2\Big)^{1/2} \\
& \kl \Big(\sum_{j=1}^m \|v_j(x)-v_j(x')\|_2^2 \|v_j(x)\|_2^2\Big)^{1/2}
+ \Big(\sum_{j=1}^m \|v_j(x)\|_2^2 \|v_j(x)-v_j(x')\|_2^2\Big)^{1/2} \\
& \kl \sup_{1\leq j\leq m} \|v_j(x)-v_j(x')\|_2 \Big\{ \Big(\sum_{j=1}^m \|v_j(x)\|_2^2\Big)^{1/2}
+ \Big(\sum_{j=1}^m \|v_j(x')\|_2^2\Big)^{1/2} \Big\} \\
& \kl 2R \sup_{1\leq j\leq m} \|v_j(x)-v_j(x')\|_2 \pl .
\end{align*}
Let $T$ denote the maximal family of elements in $D$ such that $\inf_{x \neq x' \in T} \|\beta(x)-\beta(x')\|_2>\eps$. Then it follows that $\inf_{x\neq x' \in T} \|v(x)-v(x')\|_{\ell_\infty^m(\ell_2^d)}>\frac{\eps}{2R}$. This implies that
\begin{align*}
N(\beta(D),\eps B_{\ell_2^m})
& \kl |T| \kl N\Big(v(D),\frac{\eps}{4R}B_{\ell_\infty^m(\ell_2^d)}\Big) \\
& \kl N\Big(v(\sqrt{s}K),\frac{\eps}{4R}B_{\ell_\infty^m(\ell_2^d)}\Big)
\lel N\Big(v(K),\frac{\eps}{4R\sqrt{s}}B_{\ell_\infty^m(\ell_2^d)}\Big) \pl .
\end{align*}
Using a change of variables, this implies
\begin{align*}
\int_0^{\infty}\sqrt{\ln N(\beta(D),\eps B_{\ell_2^m})} d\eps
& \kl 4R\sqrt{s} \int_0^{\infty}\sqrt{\ln N(v(K),\eps B_{\ell_\infty^m(\ell_2^d)})} d\eps \\
& \kl cR\sqrt{s} \pl \E_{2,1}(v:X\to \ell_{\infty}^m(\ell_2^d)) \pl .
\end{align*}

Next, we compute an upper bound on the second summand in the right-hand-side of \eqref{sjoerd}. By Khintchine's inequality for all $x \in D$
\begin{align*}
\Big(\ez |\langle \bm{\xi}, \beta(x)\rangle|^k\Big)^{1/k}
{} & \kl \sqrt{k} \|\beta(x)\|_2
\lel \sqrt{k} \Big(\sum_{j=1}^m \|v_j(x)\|_2^4\Big)^{1/2} \\
{} & \kl \sqrt{k} \Big(\sum_{j=1}^m \|v_j(x)\|_2^2\Big)^{1/2} \sup_{1\leq j\leq m} \|v_j(x)\|_2
\kl \sqrt{k} R \sup_{1\leq j\leq m} \|v_j(x)\|_2 \pl .
\end{align*}
Therefore
\[
\sup_{x\in D} \Big(\ez |\langle \bm{\xi}, \beta(x)\rangle|^k\Big)^{1/k}
\kl \sup_{x\in \sqrt{s}K} \sqrt{k} R \sup_{1\leq j\leq m} \|v_j(x)\|_2
\lel \sqrt{ks} R \|v:X\to \ell_2^m(\ell_2^d)\| \pl .
\]

Combining these estimates yields the assertion. \qd

\begin{cor} Suppose the hypothesis of Lemma~\ref{key}. Then
\[
\Big(\ez \sup_{x,y\in D} \Big|\sum_{j=1}^m \xi_j \langle v_j(x),v_j(y) \rangle\Big|^k\Big)^{1/k}
\kl C \sqrt{s} \pl \Big(\sup_{x\in D} \sum_{j=1}^m \|v_j(x)\|_2^2\Big)^{1/2} \pl
\Big(\E_{2,1}(v) + \sqrt{k} \|v\|\Big) \pl .
\]
\end{cor}
\begin{proof} By the polarization identity, we have
\[
\langle v_j(x),v_j(y) \rangle \lel \frac{1}{4} \sum_{l=0}^3 \mathfrak{i}^l \langle v_j(x+\mathfrak{i}^l y), v_j(x+\mathfrak{i}^l y) \rangle \pl ,
\]
where $\mathfrak{i} = \sqrt{-1}$.
Then we apply the argument for $\widetilde{D}=\bigcup_{l=0}^3 D+\mathfrak{i}^lD$.
Note that
\begin{align*}
\sum_{j=1}^m \langle v_j(x+\mathfrak{i}^ly),v_j(x+\mathfrak{i}^ly) \rangle
& = \sum_{j=1}^m \langle v_j(x),v_j(x)\rangle + (-1)^l \sum_{j=1}^m \langle v_j(y),v_j(y)\rangle \\
& + \mathfrak{i}^l \sum_{j=1}^m \langle v_j(x),v_j(y) \rangle + \mathfrak{i}^{l+1} \sum_{j=1}^m \langle v_j(y),v_j(x)\rangle \pl .
\end{align*}
Then by the Cauchy-Schwartz inequality for $x,y \in D$ we have
\begin{align*}
\Big|\sum_{j=1}^m \langle v_j(x+\mathfrak{i}^ly),v_j(x+\mathfrak{i}^ly) \rangle \Big|
\kl 2 \sum_{j=1}^m \|v_j(x)\|_2^2 + 2 \sum_{j=1}^m \|v_j(y)\|_2^2 \kl 2 R \pl ,
\end{align*}
where $R$ is defined in \eqref{eq:defR}.
Thus, the assertion follows by replacing $\sqrt{s}$ by $4\sqrt{s}$.
\qd

\begin{prop}
\label{abstractp}
Let $H$, $K$, and $K_s$ be defined as before.
Let $\delta > 0$ and $0 < \zeta < 1$.
Let $v_1,\dots,v_m$ be independent random maps from $X$ to $\ell_2^d$ satisfying \eqref{common} with $\Phi$ and \eqref{normalization}.
Let $v=(v_j)_{1\leq j\leq m}$ denote the composite map.
Suppose that
\begin{enumerate}
  \item[i)] The linear operator $\Phi$ satisfies $\|\Phi: H\to H\| \leq 1$.
  \item[ii)] The random linear operator $v: X \to \ell_{\infty}^m(\ell_2^d)$ satisfies
     \begin{equation}
     \label{eq:abstractp:cond}
     \sup_{k \in \mathbb{N}} \frac{c\sqrt{s}}{\sqrt{m}}
     \max\Big(
     (\ez \E_{2,1}(v)^{2k})^{1/2k} , \pl
     \sqrt{\ln(\zeta^{-1})} (\ez_{v}\|v\|^{2k})^{1/2k}
     \Big)
     \leq \delta
     \end{equation}
  for an absolute constant $c$.
\end{enumerate}
Then
\begin{equation}
\label{eq:abstractp:res}
\mathbb{P}\left(
\sup_{x \in K_s} \Big|\frac{1}{m} \sum_{j=1}^m \|v_j(x)\|_2^2 - \langle x,\Phi x \rangle\Big| \geq \max(\delta,\delta^2)
\right) \leq \zeta .
\end{equation}
\end{prop}

\begin{proof}
Let $Z$ denote the left-hand side of the inequality in \eqref{eq:abstractp:res}.
Let $(v_j')_{1\leq j' \leq m}$ be independent copies of $(v_j)_{1\leq j\leq m}$.
By the standard symmetrization, we have
\begin{align*}
(\ez Z^k)^{1/k} &\kl \Big(\ez \ez \sup_{x \in K_s} \Big|\frac{1}{m} \sum_{j=1}^m \|v_j(x)\|_2^2 - \|v_j'(x)\|_2^2 \Big|^k\Big)^{1/k} \\
& \kl 2 \Big(\ez \sup_{x \in K_s} \Big|\frac{1}{m} \sum_{j=1}^m \eps_j \|v_j(x)\|_2^2\Big|^k\Big)^{1/k}
\kl 2 \sqrt{\frac{\pi}{2}}
\Big( \ez \sup_{x \in K_s} \Big|\frac{1}{m} \sum_{j=1}^m \xi_j \|v_j(x)\|_2^2\Big|^k\Big)^{1/k} \pl .
\end{align*}
By conditioning on $(v_j)_{1 \leq j \leq m}$, we deduce from Lemma~\ref{key} that
\begin{align*}
\allowdisplaybreaks
(\ez Z^k)^{1/k}
&\le \frac{c_1}{m} \Big(\ez \ez_{\bm{\xi}} \sup_{x \in K_s} \Big|\sum_{j=1}^m \xi_j \|v_j(x)\|_2^2\Big|^k\Big)^{1/k} \\
&\le \frac{c_1\sqrt{s}}{m} \Big\{\ez_{v} \Big(\sup_{x \in K_s} \sum_{j=1}^m \|v_j(x)\|_2^2\Big)^{k/2} (\E_{2,1}(v)+\sqrt{k}\|v\|)^k\Big\}^{1/k} \\
&\le \frac{c_1\sqrt{s}}{m} \Big(\ez_{v} (\E_{2,1}(v)+\sqrt{k}\|v\|)^{2k} \Big)^{1/2k} \Big\{\ez_{v} \Big(\sup_{x \in K_s} \sum_{j=1}^m \|v_j(x)\|_2^2\Big)^k \Big\}^{1/2k} \\
&\le \frac{c_1\sqrt{s}}{\sqrt{m}}
\Big(\ez (\mathcal{E}_{2,1}(v)+\sqrt{k}\|v\|)^{2k}\Big)^{1/2k}
\Big\{\ez_{v} \Big(\sup_{x \in K_s} \frac{1}{m} \sum_{j=1}^m \|v_j(x)\|_2^2\Big)^k \Big\}^{1/2k} \\
&\le \frac{c_1\sqrt{s}}{\sqrt{m}} \Big(\ez_{v} (\E_{2,1}(v)+\sqrt{k}\|v\|)^{2k} \Big)^{1/2k} \\ & \cdot \Big\{\ez_{v} \Big(\sup_{x \in K_s} \frac{1}{m} \sum_{j=1}^m \|v_j(x)\|_2^2 - \langle x,\Phi x \rangle + \langle x, \Phi x \rangle\Big)^k \Big\}^{1/2k} \\
&\kl \frac{c_1\sqrt{s}}{\sqrt{m}}
\Big\{(\ez_{v}\E_{2,1}(v)^{2k})^{1/2k}+\sqrt{k}(\ez_{v}\|v\|^{2k})^{1/2k}\Big\}
\Big(1+(\ez Z^k)^{1/k}\Big)^{1/2}
\pl .
\end{align*}
Let $b$ be the factor before $(1+(\ez Z^k)^{1/k})^{1/2}$, then we have $(\ez Z^k)^{1/k} \leq \sqrt{2}(b+b^2)$.
Since $k \in \zz$ was arbitrary, a consequence of the Markov inequality \cite[Lemma~A.1]{dirksen2015tail} implies that there exists a numerical constant $c_2$ such that
\begin{align*}
Z
&\kl \frac{c_2\sqrt{s}}{\sqrt{m}} (\ez_{v}\E_{2,1}(v)^{2k})^{1/2k}
+ \frac{c_2s}{m} (\ez_{v}\E_{2,1}(v)^{2k})^{1/k} \\
& + \frac{c_2\sqrt{s}}{\sqrt{m}} \sqrt{\ln(\zeta^{-1})} (\ez_{v}\|v\|^{2k})^{1/2k}
+ \frac{c_2s}{m} \ln(\zeta^{-1}) (\ez_{v}\|v\|^{2k})^{1/k}
\end{align*}
holds with probability $1-\zeta$. The condition in \eqref{eq:abstractp:cond} implies $Z \kl \max(\delta,\delta^2)$.
\qd

\begin{proof}[Proof of Theorem~\ref{main:rv}]
Since $K$ is of entropy type $(p,\al_d)$, for every $v: X \to \ell_{\infty}^m(H)$ we have
\[
\E_{2,1}(v) \kl M_{p,\al_d}(K) m^{1/2-1/p} (1+\ln m)^{e(p)/2} \sup_{1\leq j\leq m} \al_d(v_j) \pl .
\]
Then we get
\[
(\ez \E_{2,1}(v)^{2k})^{1/2k} \kl M_{p,\al_d}(K) \Big(\ez \sup_{1 \leq j \leq m} \al_d(v_j)^{2k}\Big)^{1/2k} m^{1/2-1/p} (1+\ln m)^{e(p)/2}
\]
and hence
\[
\frac{c\sqrt{s}}{\sqrt{m}} (\ez \E_{2,1}(v)^{2k})^{1/2k} \kl \frac{c\sqrt{s} M_{p,\al_d}(K) (1+\ln m)^{e(p)/2}}{m^{1/p}} \sup_{k\in\mathbb{N}} \Big(\ez \sup_{1\leq j\leq m} \al_d(v_j)^{2k}\Big)^{1/2k} \pl .
\]
By Proposition~\ref{abstractp}, it suffices to satisfy
\[
\frac{m^{1/p}}{(1+\ln m)^{e(p)/2}} \gl \frac{c M_{p,\al_d}(K) \sqrt{s}}{\delta} \sup_{k \in \mathbb{N}} \Big(\ez \sup_{1\leq j\leq m} \al_d(v_j)^{2k}\Big)^{1/2k} \pl
\]
and
\[
m \gl \frac{cs \ln(\zeta^{-1})}{\delta^2} \sup_{k \in \mathbb{N}} (\ez_{v}\|v\|^{2k})^{1/k} \pl .
\]
\qd

\section{Complexity of sparsity models}
\label{sec:complexity}

Our generalized sparsity model is given by scaled versions of a convex set $K$. A sufficient number of measurements for the RIP is determined by the geometry of the resulting Banach space $X$. In this section, we discuss the complexity of $K$ given in terms of $M_{p,\al_d}(K)$ for various sparsity models.

\subsection{Relaxed canonical sparsity} ${\atop}$

In many applications sparsity is implicitly controlled by the $\ell_1$ norm.
A relaxed canonical sparsity model, which includes exactly sparse vectors and their approximation with small perturbation, is defined by $K = B_1^N$.
Then the corresponding Banach space is $X = \ell_1^N$.
We derive an upper bound on $M_{2,\|\cdot\|}(B_1^N)$ by using a well known application of Maurey's empirical method, which is given in the following lemma.

\begin{lemma}(Maurey's empirical method)\label{mm}  Let $m,n\in \nz$, $H$ be a Hilbert space, and $v:\ell_1^N\to \ell_{\infty}^m(H)$.
Then
\[
\sqrt{l} e_l(v) \kl c\|v\| \sqrt{(1+\ln (N/l))(1+\ln m)} \pl.
\]
In particular, for $\dim(H)=d$,
\[
\E_{2,1}(v) \kl C \sqrt{(1+\ln N)(1+\ln m)}(1+\ln m+\ln d) \|v\| \pl .
\]
\end{lemma}

\begin{proof} The first part is a direct consequence of Maurey's empirical method \cite[Proposition~2]{carl1985inequalities}. Let $p \geq 2$ and $v': \ell_1^N \to \ell_p^m(H)$, which satisfies $\|v'\| \leq m^{1/p} \|v\|$. Since $\ell_p^m(H)$ has type 2 with constant $C_1 \sqrt{p}$, it follows from \cite[Proposition~2]{carl1985inequalities} that
\[
\sqrt{l} e_l(v) \leq \sqrt{l} e_l(v') \leq C_2 \sqrt{1 + \ln (N/l)} \sqrt{p} m^{1/p} \|v\|.
\]
Choosing $p = 1 + \ln m$ yields the first assertion.
Next we prove the second assertion.
For $l < m^2 d^2$, the first assertion implies
\begin{align*}
\sum_{l=1}^{m^2d^2-1} \frac{e_l(v)}{\sqrt{l}}
& \kl \sum_{l=1}^{m^2d^2-1} \frac{c \sqrt{(1+\ln N)(1+\ln m)} \|v\|}{l} \\
& \kl C \sqrt{(1+\ln N)(1+\ln m)} (1+ \ln m + \ln d) \|v\|.
\end{align*}
For $l \geq m^2 d^2$, by the standard volume argument (\cite[Lemma~1.7]{pisier1986probabilistic}), we have $e_l(\mathrm{Id}_{\ell_{\infty}^m(H)}) \leq md/l$. Therefore,
\[
\sum_{l=m^2d^2}^{\infty} \frac{e_l(v)}{\sqrt{l}}
\leq \sum_{l=m^2d^2}^{\infty} \frac{e_l(\mathrm{Id}_{\ell_{\infty}^m(H)}) \|v\|}{\sqrt{l}}
\leq \sum_{l=m^2d^2}^{\infty} \frac{md \|v\|}{ l^{3/2} } \leq 2 \|v\| \pl .
\]
This completes the proof. \qd

\begin{prop}\label{l1} Let $\al_d(u)\lel \|u:\ell_1^N \to \ell_2^d\|$. Then
\[
M_{2,\al_d}(B_1^N) \kl C \sqrt{1+\ln N}(1+\ln d) \pl .
\]
\end{prop}

\begin{proof} According to Lemma \ref{mm} we have
\[
\sup_{k \in \mathbb{N}} \sqrt{k} e_k(v:\ell_1^N\to \ell_{\infty}^m(\ell_2^d))
\kl c \|v\| \sqrt{1+\ln N}\sqrt{1+\ln m} \pl
\]
and
\begin{align*}
\E_{2,1}(v)
& \kl c \sqrt{1+\ln N} (1+\ln dm)\sqrt{1+\ln m} \|u\| \\
& \kl C (1+\ln d)\sqrt{1+\ln N} (1+\ln m)^{3/2} \|u\| \pl .
\end{align*}
The assertion follows from the definition in \eqref{Mpal}.  \qd

\subsection{Relaxed atomic sparsity with finite dictionary}${\atop}$
\label{subsec:finite}

We say that a vector is {\em atomic $s$-sparse} if $x$ is represented as a finite linear combination of a given set of atoms, which is called a dictionary. Here we consider a special case where the dictionary is a finite set $\{x_k \pl|\pl 1 \le k\le M\}\subset B_2^N$. A relaxed atomic sparsity model is defined by the convex hull of the dictionary, i.e. $K=\absc\{x_k \pl|\pl 1 \le k\le M\}$.
As mentioned in the introduction, it is important to observe if the point $x_i$'s are in the unit sphere $\mathbb{S}^{N-1}$, then the ``sparse'' set $K_s=\sqrt{s}K \cap \mathbb{S}^{N-1}$ is no longer in a convex hull of a set with few vectors from the unit sphere.
Then the complexity of $K$ is upper bounded by the following corollary.

\begin{cor}\label{K1} Let $K=\absc\{x_j \pl|\pl 1 \le j\le M\}$ and
$\al_d(u)\lel \|u:X\to\ell_2^d\|$.
Then
\[
M_{2,\al_d}(K)\kl C\sqrt{1+\ln M}(1+\ln d) \pl .
\]
\end{cor}

\begin{proof}
Let $v=(v_j)_{1\leq j\leq m}$. Then we have
\[
\max_{1\le j\le M} \al_d(v_j) = \|v:X\to\ell_\infty^m(\ell_2^d)\| \pl.
\]
Define $Q: \ell_1^M \to X$ so that $Q(\bm{e}_k) = x_k$ for all $k=1,\dots,M$, where $\bm{e}_1,\dots,\bm{e}_M$ are standard basis vectors in $\mathbb{R}^M$. Then $\|Q: \ell_1^M\to X\| \leq 1$.
Therefore,
\[
\|v_jQ: \ell_1^M\to\ell_2^d\| \leq \|v_j:X\to\ell_2^d\| \pl, \quad \forall j = 1,\dots,M \pl .
\]
By Lemma~\ref{mm}, we have
\[
\E_{2,1}(vQ:\ell_1^M\to\ell_\infty^m(\ell_2^d))
\kl C \sqrt{(1+\ln M)(1+\ln m)}(1+\ln md) \|v:X\to\ell_\infty^m(\ell_2^d)\| \pl ,
\]
Thus Proposition~\ref{l1} applies.
Indeed, entropy numbers are surjective, i.e.
\[
e_k(vQ)\lel e_k(v:X\to Y)
\]
for any Banach space $Y$.
Therefore, the estimate in Corollary~\ref{K1} is as tight as that in Lemma~\ref{l1}.
\qd

\subsection{Norm dual of type-$p$ Banach spaces} $\atop$
\label{subsec:dualtypep}

Next we consider the scenario where the norm dual $X^*$ has type $p$.
Let us recall that a Banach space $X$ has {\em type $p$} if there exists a constant $C$ such that for all finite sequence $(x_j)$ in $X$
\begin{equation}
\label{def:typep}
\Big(\ez \Big\|\sum_j \eps_j x_j\Big\|_X^p\Big)^{1/p} \kl C \Big(\sum_j \|x_j\|_X^p\Big)^{1/p} \pl ,
\end{equation}
where $(\eps_j)$ is a Rademacher sequence \cite{pisier1999volume}.
The type $p$ constant of $X$, denoted by $T_p(X)$, is the smallest constant $C$ that satisfies \eqref{def:typep}.
Let $\al_d$ be the operator norm. Then Maurey's method implies that $K$ has type $(p,\al_d)$ if $X^*$ has type $p$, which will be shown using the following lemma.

\begin{lemma}\label{eduality} Let $X$ be a Banach space such that $X^*$ has type $1<p<2$ and $v:X\to \ell_{\infty}^m$. Then
\[
\sum_{k=1}^\infty \frac{e_k(v)}{\sqrt{k}}
\kl c(p) T_p(X^*)^{p'+1} \|v\| m^{1/p-1/2} (1+\ln m)^{1/2} \pl ,
\]
where $c(p)$ is a constant that only depends on $p$.
Moreover, for $p=2$,
\[
\sum_{k=1}^\infty \frac{e_k(v)}{\sqrt{k}}
\kl c T_2(X^*)^3 \|v\| (1+\ln m)^{3/2} \pl
\]
for a numerical constant $c$.
\end{lemma}

The following corollary is a direct consequence of Lemma~\ref{eduality}.
\begin{cor}
\label{mpal_dualtypep}
By definition, if $X^*$ is of type $p$ for $1<p<2$, then
\[
M_{p,\|\cdot\|}(K) \leq c(p) T_p(X^*)^{p'+1} m^{2/p-1} (1+\ln m)^{1/2} \pl .
\]
For $p=2$,
\[
M_{2,\|\cdot\|}(K) \leq c T_2(X^*)^3 (1+\ln m)^{3/2} \pl .
\]
\end{cor}

\begin{proof}[Proof of Lemma~\ref{eduality}]
Let $v^*:\ell_1^m\to X^*$ denote the adjoint of $v$.
Then Maurey's empirical method \cite{carl1985inequalities} implies
\[
e_k(v^*) \kl c T_p(X^*) \|v\| \left(\frac{\ln m}{k}\right)^{1/p'} \pl .
\]
Moreover the duality of entropy numbers \cite[Proposition 4 and Lemma C]{bourgain1989duality} implies that for all $l \in \mathbb{N}$
\[
\sum_{k=1}^l e_k(v) \kl (CT_p(X^*))^{p'} \sum_{k=1}^l e_k(v^*)
\]
for a numerical constant $C$. Using Hardy's inequalities (e.g., \cite{pietsch1980weyl}), this implies \begin{align*}
\sum_{k=1}^l \frac{e_k(v)}{\sqrt{k}}
&\kl (CT_p(X^*))^{p'} \sum_{k=1}^l k^{-1/2}e_k(v^*) \\
&\kl (CT_p(X^*))^{p'+1}\|v\| (1+\ln m)^{1/p'} \sum_{k=1}^l k^{-1/2-1/p'}  \\
&\kl c \Big(\frac{1}{2}-\frac{1}{p'}\Big)^{-1} (CT_p(X^*))^{p'+1}\|v\| (1+\ln m)^{1/p'} l^{1/2-1/p'}  \pl .
\end{align*}
On the other hand, by the standard volume argument, we have $e_k(\mathrm{Id}_{\ell_1^m})\kl e^{-ck/m}$. Now we use the fact that
\[
\int_b^\infty x^{-1/2}e^{-x/a} dx
\lel a^{1/2} \int_{b/a}^\infty y^{-1/2}e^{-y} dy
\kl c a b^{-1/2} e^{-b/a}
\]
holds for $b/a\gl 2$. Therefore $c l/m \gl 2$ implies
\[
\sum_{k\gl l} k^{-1/2}e^{-ck/m} \kl \frac{m}{l^{1/2}} e^{-cl/m} \pl .
\]
Thus for $l=m\sqrt{\ln m}$ this is bounded by a constant. Thus we deduce from $e_k(v)\kl \|v\|e_k(\mathrm{Id}_{\ell_{\infty}^N})$ that
\begin{align*}
\sum_{j=1}^{\infty} \frac{e_k(v)}{\sqrt{k}}
\kl c \left(\frac{1}{2}-\frac{1}{p'}\right)^{-1} (CT_p(X^*))^{p'+1}\|v\| m^{1/2-1/p'} (1+\ln m)^{1/2-1/p'} (1+\ln m)^{1/p'} \pl .
\end{align*}
Let $c(p) = c(1/2-1/p')^{-1} C^{p'}$. This proves the first part.
For $p=2$, we let $l=\ln m+\ln\ln m \kl 2\ln m$ in splitting $\sum_{k=1}^\infty e_k(v^*)/\sqrt{k}$ into two partial summations. Moreover, $T_2(X^*)^3=T_2(X^*)T_2(X^*)^2$ comes from the duality of entropy numbers and Maurey's method.
\qd

\subsection{Unconditional basis and lattices} ${\atop}$

Let us first consider the case where $X$ is a finite dimensional Banach lattice.
Let us recall that the unit ball $K$ of a lattice $X$ in $\rz^N$ is a convex symmetric set such that
\begin{equation}\label{eps}
D_{\eps}(K)\lel K \pl
\end{equation}
holds for all $\eps \in \{-1,1\}^N$, where $D_\eps$ is a diagonal operator that performs the element-wise multiplication with $\eps$. In the complex case we require this condition for all $\eps\in \mathbb{T}^N$, where $\mathbb{T}$ denotes the set of unit modulus complex numbers. Equivalently, the norm $\|\cdot\|_X$ given as the Minkowski functional of $K$ satisfies
\[
\|(x_i)_{1\le i\le N}\|_X \lel \| (|x_i|_{1\le i\le N})\|_X \pl .
\]

\begin{rem}{\rm
\label{inf_lattice}
The abstract definition of a Banach lattice is more involved. In practical purpose, we may always assume that a lattice is given by a norm on measurable function $f$ on $(0,\infty)$ or $[0,1]$ such that
\[
\||f|\|_X = \|f\|_X,
\]
where $|f|$ is defined by $|f|(t) = |f(t)|$ for all $t$ in the domain of $f$. (See \cite{lindenstrauss1996classical} for more details.)
In this setup, all arguments in this section also apply to infinite dimensional Banach lattices.}
\end{rem}

For $k\in\nz$, let $\gamma_{\infty}^k$ denote the homogeneous function on the maps $u:X \to \ell_2^d$ defined by
\[
\gamma_{\infty}^k(u) \lel \inf_{u\lel ab} \|b:X\to \ell_{\infty}^k\| \pl \|a:\ell_{\infty}^k\to \ell_2^d\| \pl .
\]
Note that $\gamma_{\infty}^k$ is not necessarily a valid norm. The following well-known Lemma is crucial in our context (see \cite{pisier1986factorization}).

\begin{lemma}\label{ggn} Let $X$ be a lattice as above and $u:X\to \ell_2^d$. Then
\[
\gamma_{\infty}^{2d}(u) \kl C \Big\|\Big(\sum_{j=1}^d |u^*(\bm{e}_j)|^2\Big)^{1/2}\Big\|_{X^*} \kl 2C \ell(u(K)) \pl .
\]
\end{lemma}

\begin{proof}
There exists an isomorphic embedding $a: \ell_2^d \hookrightarrow \ell_1^{2d}$ due to Kashin (see e.g., \cite{pisier1999volume}). Since $X$ is a lattice, it follows that $X(\ell_2^d)$ is also embedded into $X(\ell_1^{2d})$. Moreover, since $\ell_2^d$ and $\ell_1^{2d}$ are finite dimensional lattices, the dual spaces of $X(\ell_2^d)$ and $X(\ell_1^{2d})$ are given by $X^*(\ell_2^d)$ and $X^*(\ell_\infty^{2d})$, respectively \cite{lindenstrauss1996classical}.
By the Hahn-Banach theorem, for every $x^*=(x_j^*)_{1\leq j\leq d} \in X^*(\ell_2^d)$ with $\|x^*\|_{X^*(\ell_2^d)} = 1$, there exists $y^*=(y_i^*)_{1\leq i\leq 2d} \in X^*(\ell_\infty^{2d})$ that satisfies $\|y^*\|_{X^*(\ell_\infty^{2d})} \leq C_1$ for a numerical constant $C_1$ and $x^* = (\mathrm{Id} \otimes a^*) y^*$, i.e.
\[ x_j^* \lel \sum_{i=1}^{2d} a_{ij} y_i^*, \quad j = 1,\dots,d \pl . \]

Choose $x^* = (x_j^*)_{1\leq j\leq d} = (u^*(\bm{e}_j))_{1\leq j\leq d}$. Then let $y^* = (y_i^*)_{1\leq i\leq 2d}$ be the extension of $x^*$ as above. Define a linear operator $\phi: X\to \ell_\infty^{2d}$ by $\phi(x) = (\langle y_i^*, x\rangle)_{1\leq i\leq 2d}$ for $x \in X$. Then $u$ is factorized as $u = a^*\phi$.
Since $X^*(\ell_\infty^{2d}) \subset \ell_\infty^{2d}(X^*)$, the operator norm of $\phi$ is upper bounded by
\[
\|\phi: X\to\ell_\infty^{2d}\| = \sup_{1\leq i\leq 2d} \|y_i^*\|_{X^*}
= \|y^*\|_{\ell_\infty^{2d}(X^*)}
\kl \|y^*\|_{X^*(\ell_\infty^{2d})}
\kl C_1 \|x^*\|_{X^*(\ell_2^d)} \pl.
\]
Therefore,
\[
\gamma_{\infty}^{2d}(u)\kl C_1 \|x^*\|_{X^*(\ell_2^d)} \|a^*: \ell_\infty^{2d}\to \ell_2^d\| \pl.
\]
Here $\|a^*: \ell_\infty^{2d}\to \ell_2^d\|$ is a numerical constant.
Note that $\|x^*\|_{X^*(\ell_2^d)}$ is written as
\begin{equation}
\label{XKell2norm}
\|x^*\|_{X^*(\ell_2^d)}
\lel \Big\|(u^*(\bm{e}_j))_{j=1}^d\Big\|_{X^*(\ell_2^d)}
\lel \Big\|\Big(\sum_{j=1}^d |u^*(\bm{e}_j)|^2\Big)^{1/2}\Big\|_{X^*} \pl .
\end{equation}
This proves the first assertion.
Finally, by Khintchine's inequality, the last term in \eqref{XKell2norm} is upper bounded by
\begin{align*}
\Big\|\Big(\sum_{j=1}^d |u^*(\bm{e}_j)|^2\Big)^{1/2}\Big\|_{X^*}
& \kl \sqrt{2} \Big\| \ez \Big|\sum_{j=1}^d \xi_ju^*(\bm{e}_j)\Big| \Big\|_{X^*}
\kl \sqrt{2} \ez \Big\|\pl \Big|\sum_{j=1}^d \xi_ju^*(\bm{e}_j)\Big|\pl \Big\|_{X^*} \\
& \lel \sqrt{2} \ez \Big\|\sum_{j=1}^d \xi_ju^*(\bm{e}_j)\Big\|_{X^*}
\lel \sqrt{2} \ez \sup_{x \in K} \Big\langle \sum_{j=1}^d \xi_ju^*(\bm{e}_j), x \Big\rangle \\
& \lel \sqrt{2} \ez \sup_{x \in K} \Big\langle \sum_{j=1}^d \xi_j\bm{e}_j, u(x) \Big\rangle
\lel \sqrt{2} \ell(u(K)) \pl ,
\end{align*}
where $\xi_1,\dots,\xi_d$ are independent copies of $\xi \sim \mathcal{N}(0,1)$.
\qd

The above lemma suggests that a good choice for $\al_d$ is given by
\[
\al_d(u)= \Big\| \Big(\sum_{j=1}^d |u^*(\bm{e}_j)|^2 \Big)^{1/2} \Big\|_{X^*} \pl .
\]
This definition works verbatim for Banach lattices.

\begin{theorem}\label{unc}
Let $K$ be a convex symmetric set and $1< p\le 2$.
Suppose that $K$ satisfies \eqref{eps}. Then
\[
M_{p,\al_d}(K) \kl c(p) T_p(X^*)^{p'+1} d^{1/p-1/2}(1+\ln d)^{e(p)/2} \pl .
\]
\end{theorem}

\begin{proof} Let $(v_j)_{j=1}^m$ be maps with $\al_d(v_j)\le 1$. Then we find a factorization $v_j=a_jb_j$ with $\|b_j:X\to \ell_{\infty}^{2d}\|\le 1$ and $\|a_j:\ell_{\infty}^{2d}\to \ell_2^d\|\kl C$. This allows us to define $b=(b_j):X\to \ell_{\infty}^{2md}$ given by the blocks. Similarly, we may define the block diagonal map $a=(a_j):\ell_{\infty}^m(\ell_{\infty}^{2d})\to \ell_{\infty}^m(\ell_2^d)$ of norm $\le C$. According to Lemma~\ref{eduality}, we have
\[
\E_{2,1}(b)\kl c(p) T_p(X^*)^{p'+1} (2md)^{1/p-1/2} (1+\ln (md))^{e(p)/2} \pl .
\]
Using $v=ab$ implies the assertion.\qd

\subsection{Schatten classes} ${\atop}$

Schatten $p$-classes are examples of noncommutative $L_p$ spaces. In that sense, we should expect that the results from the previous section extend to those ``noncommutative lattices''.
In this context the maps $v$ which extract a row or a column from the matrix are canonical, although not very clever choices, for Hilbert space valued measurements.
Let us also note that just using the operator norm $\|v: S_p^n \to \ell_2^d\|$ is a ``bad'' idea. The column and row projection certainly do not admit any entropy decay.

Before we start estimating the relevant constants, we should mention that for $1\le q\le 2$ the unit ball $B_{S_q^{n_1,n_2}}$ satisfies
\[
B_{S_q^{n_1,n_2}} \subset B_2^N \pl,
\]
where $N = n_1 n_2$. Let us also observe that $K=B_{S_q^{n_1,n_2}}$ has enough symmetries. Indeed, we have affine isotropic actions of $\zz_{n_1^2}$ and $\zz_{n_2^2}$ respectively as left and right multiplications, and the argument for tensor products in Section~\ref{sec:tensor} also applies here and shows the commuting action of $\zz_{n_1^2}\times \zz_{n_2^2}$ is also isotropic. An analogue of Lemma~\ref{ggn} is stated for $X = S_q$. For simplicity of notation, we state our results in the square case ($n_1=n_2=n$). The rectangular case can be shown in a similar way.

\begin{lemma}
\label{spn} Let $1\le q\le 2$ and $u:S_q\to \ell_2^d$. Then
\[
\gamma_{\infty}^{cd^2}(u) \kl C \Big(\Big\|\Big(\sum_{j=1}^d |u^*(\bm{e}_j)|^2\Big)^{1/2}\Big\|_{q'} + \Big\|\Big(\sum_{j=1}^d |u^*(\bm{e}_j)^*|^2\Big)^{1/2}\Big\|_{q'} \Big) \pl .
\]
\end{lemma}

\begin{proof} Let us denote by $G_q^d(X)=\{\sum_{j=1}^d \xi_j \ten x_j\} \subset L_q(\Om,P;X)$ the span of the gaussian vectors and by $G_q^d(X)^*$ the dual space. In \cite[Theorem~1.13]{pisier2009remarks} Pisier proved that for $1\le q < 2$ the space $G_q^d\subset L_q$ completely embeds into $\ell_q^{cd^2}$ for some universal constant $c$. This implies that there exists a map $a:G_q^d\hookrightarrow \ell_q^{cd^2}$ which is an $2$-isomorphism on its image so that $G_q^d(S_q) \hookrightarrow \ell_q^{cd^2}(S_q)$.
By duality we find that $G_q^d(S_q)^*$ is a quotient of $\ell_{q'}^{cd^2}(S_{q'})$.
Therefore, for every $x^* = (x_j^*)_{1\leq j \leq d} \in G_q^d(S_q)^*$ with $\|x^*\|_{G_q^d(S_q)^*} = 1$, there exists $y^* = (y_i^*)_{1\leq i\leq cd^2} \in \ell_{q'}^{cd^2}(S_{q'})$ that satisfies $\|y^*\|_{\ell_{q'}^{cd^2}(S_{q'})} \leq C_1$ for a numerical constant $C_1$ and $x^* = (\mathrm{Id} \otimes a^*) y^*$. Note that there exists an isometric isomorphism $\iota:\ell_2^d\to (G_q^d)^*$. We pick $x^* = (x_j^*)_{1\leq j\leq d} = (u^*\iota^*\iota(\bm{e}_j))_{1\leq j\leq d} = (u^*(\bm{e}_j))_{1\leq j\leq d}$ and let $y^*$ be the image of $x^*$ via $\mathrm{Id} \otimes a^*$ as above. Define $\phi:S_q\to\ell_{q'}^{cd^2}$ by $\phi(x) = (\langle y_i^*, x\rangle)_{1\leq i\leq cd^2}$ for $x \in S_q$. Let $b:S_q\to\ell_\infty^{cd^2}$ be a linear operator given by $b(x) = (\langle y_i^*, x \rangle / \|y_i^*\|_{S_{q'}})_{1\leq i\leq cd^2}$. Let $D_\si:\ell_\infty^{cd^2}\to\ell_{q'}^{cd^2}$ be a diagonal operator given by $\sigma = (\|y_i^*\|_{S_{q'}})_{1\leq i\leq cd^2} \in \ell_{q'}^{cd^2}$. Then $\|b\| \leq 1$ and $\|\phi\| = \|D_\si\| = \|\si\|_{q'}$. In other words, $\phi$ is factorized as
$\phi = D_\si b$
so that
\[
\|b\| \|\si\|_{q'} \kl \|\phi\|
\lel \|y^*\|_{\ell_{q'}^{cd^2}(S_{q'})}
\kl C_1 \|x^*\|_{G_q^d(S_q)^*}
\pl .
\]
This implies $u$ satisfies $\gamma_{\infty}^{cd^2}(u)\kl C_1 \|(u^*(\bm{e}_j)_{1\leq j\leq d}\|_{G_q^d(S_q)^*} \|a^*\|$. Note that $\|a^*\|$ is a constant. The dual space of $G_q^d(S_q)$ is well-understood via noncommutative Khintchine inequalities (see \cite{pisier2003introduction}) and the norm is given by the right-hand-side of our assertion for $1\le q\le 2$.\qd

\begin{rem}{\rm For $q\gl 2$, we expect a similar result using polynomial $\La$-cb sets. We leave the details to a future publication.}
\end{rem}

\begin{cor}\label{sq} Let $1\le q\le 2$ and
\[
\al_d(u) \lel \Big\|\Big(\sum_{j=1}^d |u^*(\bm{e}_j)|^2\Big)^{1/2}\Big\|_{q'} + \Big\|\Big(\sum_{j=1}^d |u^*(\bm{e}_j)^*|^2\Big)^{1/2}\Big\|_{q'}
\]
be given by the noncommutative square function. Let  $K_q=B_{S_q^n}\subset B_2^{n^2}$. Then
\[
M_{2,\al_d}(K_q)\kl C (1+\ln d)^{3/2} (q')^{3/2} \pl
\]
for $q>1$ and $M_{2,\al_d}(K_1)\kl (1+\ln d)^{3/2} (1+\ln n)^{3/2}$.
\end{cor}

\begin{proof} For $v=(v_j)$, we deduce from Lemma \ref{eduality} that
\begin{align*}
\E_{2,1}(v)
& \kl c(T_2(S_{q'}))^3 (1+\ln mcd^2)^{3/2} \sup_j \al_d(v_j) \\
& \kl c(\sqrt{q'})^3 (1+\ln d)^{3/2} (1+\ln m)^{3/2} \sup_j \al_d(v_j) \pl .
\end{align*}
In fact this estimate is rather rough, it would be better to work with the maximum of $(1+\ln m)^{3/2}$ and $(1+\ln d)^{3/2}$.  \qd

\begin{rem} {\rm We have shown in Section~\ref{subsec:finite} that convex hulls generated by few points induce sparsity models. Our estimate above provides a noncommutative analogue as follows: Fix $k\in \nz$ and let $Q:S_1^k\to \ell_2^N$ such that
\[
\Big\|\Big(\sum_{j=1}^N |Q^*(\bm{e}_j)|^2\Big)^{1/2}\Big\|_{M_k}
+\Big\|\Big(\sum_{j=1}^N |Q^*(\bm{e}_j)^*|^2\Big)^{1/2}\Big\|_{M_k} \kl 1 \pl .
\]
This replaces the condition $x_i\in B_2^N$ for the commutative space.  Let $K=Q(B_{S_1^k})$. Then we deduce from Corollary \ref{sq} that
\[
M_{2,\al_d}(K)\kl C (1+\ln d)^{3/2} (1+\ln k)^{3/2} \pl .
\]
Of course such a set $K$ need not necessarily have enough symmetries. Let $G$ be a finite group. Then we may consider the map $Q^G:\ell_1^{|G|}(S_1^{k}) \to \ell_2^N$ given by
\[
Q^G((x_g)) \lel \sum_{g \in G} Q(x_g)
\]
and the larger body $K^G=Q^G(B_{\ell_1^{|G|}(S_1^k)})$. Since $\ell_1^{|G|}(S_1^k)$ is isometrically embedded into $S_1^{k|G|}$, our estimate also implies
\[
M_{2,\al_d}(K^G)\kl C (1+\ln d)^{3/2} (1+\ln k + \ln |G| )^{3/2} \pl .
\]
Hence, up to a logarithmic factor of a higher order, the set convex hull $K^G$ of the $G$-orbit of $K$ satisfies the same estimates as convex combinations of few points. Note however that for a map $Q:\ell_1^k\to \ell_2^N$ we have
\[
\|Q\|
\lel \sup_j \|Q(\bm{e}_j)\|_2
\lel \Big\|\Big(\sum_{j=1}^N |Q^*(\bm{e}_j)|^2\Big)^{1/2}\Big\|_{\infty} \pl,
\]
and hence our noncommutative condition is a suitable generalization of the commutative case.}
\end{rem}

\section{Sparsity models with enough symmetries}
\label{sec:symmetry}

\subsection{Banach spaces with enough symmetries}$\atop$

In many physical situations, one considers particular isometries groups for the underlying configuration space. In the Banach space literature, a space $X$ is called to have {\em enough symmetries} if there is an affine representation $\si:G\to L(X)$ such that
\begin{enumerate}
  \item[i)] $\|\si(g)x\|_X=\|x\|_X$;
  \item[ii)] $\si(g)T=T\si(g), \quad \forall g \in G \implies T = \la \mathrm{Id}$.
\end{enumerate}

In this section we will assume that $X$ is finite dimensional and  $G$ is compact. Let us recall that an affine representation is almost multiplicative, i.e. there exists $\phi: G \times G \to [0,2\pi)$ such that
\[
\si(g)\si(h) \lel e^{\mathfrak{i}\phi(g,h)}\si(gh) \pl, \quad \forall g,h \in G \pl.
\]
These representations are usually obtained from the representation of the Lie algebra. Affine representations yield an honest group presentations $\pi:G\to L(X)$  by conjugation so that
\[ \pi(g)(T) \lel \si(g)T\si(g^{-1}) \pl, \quad \forall g \in G, \pl T \in L(X) \pl .\]
Indeed, we have
 \begin{align*}
  &\pi(g)\pi(h)T
 \lel \si(g)\si(h)T\si(h)^{-1}\si(g)^{-1} \\
  &= \si(gh) e^{\mathfrak{i}\phi(gh)} Te^{-\mathfrak{i}\phi(gh)}\si(h^{-1}g^{-1}) \lel \pi(gh)T \pl .
   \end{align*}

The next result is well-known (see e.g. \cite{pisier1999volume,tomczak1989banach}), we include a sketch of the proof for convenience.

\begin{lemma}\label{sym} Let $X$ be a Banach space of dimension $N$ with enough symmetries with respect to a compact group $G$. Then there exists an inner product on $X$, with corresponding Hilbert space $H$,  and an affine representation $\si:G\to U(H)$, where $U(H)$ denote the group of unitary operators on the corresponding Hilbert space $H$, such that
\begin{equation}\label{symm}
\int_G \si(g)T\si(g^{-1}) d\mu(g) \lel \frac{\mathrm{tr}(T)}{N} \mathrm{Id} \pl .
\end{equation}
\end{lemma}

\begin{proof}
Let $\al$ be an ideal norm on $L(\ell_2^N,X)$ and $u_0:\ell_2^N\to X$ be the Lewis map with respect to $\al$ such that
\begin{equation} \label{oo}
\det(u_0) \lel \sup_{\al(u)\le 1}|\det(u)| \pl .
\end{equation}
Here we have chosen a fixed basis on $X\cong \kz^N$ in order to calculate the determinant. Note that
\[
\si(g)(B_X)\lel B_X
\]
implies $\mathrm{vol}(\si(g)B_X)=|\det(\si(g))|^N \mathrm{vol}(B_X)$ and hence $|\det(\si(g))|=1$. Thus $\si(g)u_0$ also attains the maximum in \eqref{oo}. Since the ellipsoid $\E=u_0(B_2^N)$ is unique for any tensor norm on $L(\ell_2^N,X)$, see \cite{pisier1999volume}, we deduce that $\si(g)(\E)=\E$. This implies that $u_0^{-1}\si(g)u_0(B_2^N)\subset B_2^N$ and hence $\hat{\si}(g)=u_0^{-1}\si(g)u_0$ is a contraction in $L(\ell_2^N)$. Applying this to $\si(g^{-1})$, which differs from $\si(g)^{-1}$, up to a scalar of absolute value $1$, we deduce that $\hat{\si}(g)$ is a unitary, i.e. $\hat{\si}(g)B_2^N=B_2^N$. The Hilbert space $H$ is now obtained from the norm $\|x\|_H=\|u_0^{-1}(x)\|_2$, and then $\si(g)$ simultaneously preserves the unit ball of $X$ and is a unitary on $H$.
In particular, the linear map
\[
\Phi(T) \lel \int_G \si(g)T\si(g^{-1}) d\mu(g)
\]
satisfies $\si(g_0)\Phi(T)\si(g_0^{-1})=\Phi(T)$ for all $g_0$. Thus $\Phi(T)=\la(T)\mathrm{Id}$, where
\[
\la(T)N \lel \mathrm{tr}(\Phi(T)) \lel \int_G \mathrm{tr}(\si(g)T\si(g^{-1})) d\mu(g) \lel \mathrm{tr}(T) \pl .
\]
The assertion follows by normalization. \qd

We see that equivalently an $N$-dimensional Banach space $X$ with enough symmetries is given by $(\cz^N,\|\cdot\|_X,\|\cdot\|_H)$ and an affine representation $\si:G\to O_N$ such that $\si(g)$ is an isometry on $X$ and on $H$ \emph{simultaneously}.

\subsection{Examples of isotropic affine representations}$\atop$

Let $G$ be a compact group and $\si:G\to O_N$ be an irreducible representation. Then $\si(G)'=\cz \mathrm{Id}$ show that $(G,\si)$ is isotropic.
Therefore, even for finite groups, it is impractical to provide an exhaustive list of affine isotropic representations.
For concreteness, we provide a few examples in the following.

\subsubsection{Quantum Fourier transform}

The smallest group  we are aware of is the group $\zz_N^2$. Let $\Sh(\bm{e}_r)=\bm{e}_{r+1}$ be the cyclic shift and $\La(\bm{e}_r)=e^{\frac{\mathfrak{i} 2\pi r}{N}} \bm{e}_r$ be the diagonal operator representing the modulation with the $N$-th primitive root of unity. Then $\si:\zz_N^2\to O_N$ defined by
\[
\si(l,k) \lel \La^l\Sh^k, \quad \forall (l,k) \in \zz_N^2
\]
is an affine representation, because
\[
\si(l,k)\si(l',k')
\lel \La^l\Sh^k\La^{l'}\Sh^{-k} \Sh^{k+k'}
\lel e^{\frac{-\mathfrak{i} 2\pi l'k}{N}}\si(l+l',k+k') \pl
\]
satisfies the usual Heisenberg relations. Since $\La$ is a single generator of the $C^*$-algebra $\ell_{\infty}^N$, we see that every matrix $a$ commuting with $\La$ is a diagonal matrix. Every diagonal matrix commuting with the shift has constant entries. Thus the commutant of the group actions is given by $(\si(\zz_{N^2}))'=\cz \mathrm{Id}$.

\subsubsection{Random sign and shifts}

Consider an affine representation $\tilde{\si}$ of $\widetilde{G}=\{-1,1\}^N\rtimes \zz_N$ via diagonal matrices and shift matrices.

Let us show that we deal in fact with a suitable representation. Indeed, let $\mathrm{Sh}(\bm{e}_r)=\bm{e}_{r+1}$. Then
 \[ \textrm{Sh}^{-1}D_{\eps}\mathrm{Sh}\lel D_{\eps'} \]
with entries $\eps'_r=\eps_{r+1}$. This means that
\[
\tilde{\si}(\widetilde{G}) \lel \Big\{ D_{\eps}\mathrm{Sh}^k \pl \Big| \pl \eps\in \{-1,1\}^N, k\in \zz_N\Big\} \subset O_N
\]
is a subgroup and the normalized counting measure is the Haar measure. $\widetilde{G}$ is indeed the semi-direct product, and it easily checked that $(\tilde{\sigma}\widetilde{G})'=\cz \mathrm{Id}$.

\subsubsection{Clifford group and Schatten class}$\atop$

For the Schatten class $X=S_1^n$ we have several interesting choices of group actions. Indeed let us assume that $\si:G\to U_n$ is an affine isotropic action, then
$\tilde{\si}: G\times G\to U(S_2^n)$ given by
\[
\tilde{\si}(g,g')(x)\lel \si(g)x\si(g')^*
\]
also defines an isotropic action on $S_2^n$.

\begin{exam} For our quantum unitaries $\si(k,j)=\La^k{\rm Sh}^j$ we see that the product is again of the same form so that we may effectively sample with the quantum Fourier transform matrices as measurement.
\end{exam}

\begin{exam} For $n=2^k$, we may use the standard Clifford generators \cite{pisier2003introduction}
\[
c_rc_s+c_sc_r\lel \delta_{rs} 1 \pl ,\pl c_r^2=1\pl c_r=c_r^* \pl .
\]
Then we define affine representation $\si:\{0,1\}^k \to U_n$ via
\[
\si(\eps) \lel \prod_{\eps_k=1} c_k \pl .
\]
Here the product is ordered. Note that $\si(\eps)\si(\eps')=\pm \si(\eps\eps')$. It is easy to check that this action is isotropic.
\end{exam}

\begin{exam} For $n=2$, we may use also the Pauli matrices $I,\eps,J,\mathfrak{i}\eps J$ given by
\[
I \lel \kla \begin{array}{cc} 1& 0\\0&1\end{array} \mer, \quad
\eps \lel \kla \begin{array}{cc} 1& 0\\0&-1\end{array} \mer,\quad
J \lel \kla \begin{array}{cc} 0& 1\\ 1& 0\end{array} \mer \pl .\]
Then we find an affine representation $\si_2$ of $\zz_2^2$ on $M_2$, and the tensor product $\si_2^{k}:\zz_2^{2k}\to M_2^{\ten_k}=M_{2^k}$ gives exactly the measurements described by Liu \cite{liu2011universal}.
\end{exam}

\section{RIP via group action}
\label{sec:grmeas}

In this section, we illustrate implication of the general RIP result in Theorem~\ref{main:rv} for prototype sparsity models when $K$ has enough symmetries with an affine representation and the measurement operator is group structured accordingly.

\subsection{Relaxed atomic sparsity with a finite dictionary}$\atop$

First, we consider the case where $K$ is a polytope given as an absolute convex hull of finitely many points. The corresponding sparsity model is is a generalization of the canonical sparsity model with the unit $\ell_1$ ball.

\begin{theorem}[Polytope]
\label{randrv}
Under the hypothesis of Theorem~\ref{main:rv}, suppose in addition that $H = \ell_2^N$ and $K$ is an absolute convex hull of $M$ points, and $K$ has enough symmetry with an isotropic affine representation $\sigma: G \to O_N$. Let $g$ be a Haar-distributed random variable on $G$ and $g_1,\dots,g_m$ be independent copies of $g$. Let $u:X\to\ell_2^d$ satisfy $\mathrm{tr}(u^*u) = N$. Then there exists a numerical constant $c$ such that $A = \frac{1}{\sqrt{m}} (u\si(g_j))_{1\leq j \leq m}$ satisfies the RIP on $\Gamma_s$ with constant $\delta$ with probability $1-\zeta$ provided
\[
m \geq
c \delta^{-2} s \max((1+\ln m)(1+\ln md)^2 (1+\ln M), \ln(\zeta^{-1})) \|u:X\to\ell_2^d\|^2 \pl .
\]
\end{theorem}

\begin{proof}
By the isotropy, we have $\ez \si(g)^*u^*u\si(g) = \mathrm{Id}$.
Let $\al_d$ be the operator norm. Then, since $K$ is $G$-invariant,
\[
\al_d(u\si(g_j)) = \|u\si(g_j):X\to\ell_2^d\| = \|u:X\to\ell_2^d\| \pl.
\]
Note that $\|u:X\to\ell_2^d\|$ is no longer random.
The assertion follows by applying the above results with the upper bound on $M_{2,\|\cdot\|}(K)$ by Lemma~\ref{K1} to Remark~\ref{rem:main:rv}.
\qd

\begin{rem}{\rm
When $K$ is not $G$ invariant, one can show the RIP for
\[
\widetilde{K} = \absc\{\si(g)x_j \pl|\pl 1\le j\le M, \pl g \in G\}
\]
instead of $K$. By construction, $\widetilde{K}$ is $G$ invariant. Moreover, since $K \subset \widetilde{K}$, it follows that $\|x\|_{\widetilde{X}} \leq \|x\|_X$ for all $x \in X$, where $\widetilde{X}$ is the Banach space with unit ball $\widetilde{K}$. Therefore the RIP on $\Gamma_s(\widetilde{K})$ implies the RIP on $\Gamma_s$. For example, if $G = \zz_N^2$, then this replacement of $K$ by $\widetilde{K}$ will increase the number of measurements for the RIP in Theorem~\ref{randrv} by an additive term of $O(\ln N)$.}
\end{rem}

In a special case where $K=B_1^N$ and $d=1$, there is $\eta\in X^*$ such that the map $v$ is given by $u(x) = \langle\eta,x\rangle$ for $x \in X$ and $\|u:X\to\ell_2^d\|$ reduces to $\|\eta\|_\infty$. In particular, for $\eta = [1,\dots,1]^\top \in\rz^N$, we have $\mathrm{tr}(u^*u) = \|\eta\|_2^2 = N$ and Theorem~\ref{randrv} reproduces the known RIP result for a partial Fourier operator on the canonical sparsity model \cite{rudelson2008sparse,rauhut2010compressive,dirksen2015tail}.

\subsection{Dual of type-$p$ Banach spaces}$\atop$

Next we consider the case where the norm dual $X^*$ has type $p$ and the measurements are scalar valued ($d=1$).
Let $\al_d$ be the operator norm. Then Lemma~\ref{eduality} implies that $K$ has type $(p,\al_d)$ if $X^*$ has type $p$. Therefore we get the following theorem.

\begin{theorem}[Dual of type $p$]\label{point}
Let $X$ be an $N$-dimensional Banach space such that i) $X^*$ has type $1< p\le 2$; ii) the unit ball $K$ has enough symmetry given by $(\cz^N,\|\cdot\|_X,\|\cdot\|_2)$ and an affine representation $\sigma: G \to O_N$. Let $g$ be a Haar-distributed random variable on $G$ and $g_1,\dots,g_m$ be independent copies of $g$.
Let $\eta\in\cz^N$ such that $\|\eta\|_2=\sqrt{N}$.
Then
\[
\mathbb{P}\left(
\sup_{\|x\| \leq \sqrt{s},~ \|x\|_2=1} \Big|\frac{1}{m}\sum_{j=1}^m |\langle \eta, \si(g_j)x\rangle|^2 - \|x\|_2^2\Big| \geq \max(\delta,\delta^2)
\right) \leq \zeta
\]
provided
\begin{align*}
\frac{m^{1-1/p}}{(1+\ln m)^{e(p)/2}}
&\geq c_p \delta^{-1} \sqrt{s} T_p(X^*)^{p'+1} \|\eta\|_{X^*} \pl
\intertext{and}
m &\gl c \delta^{-2} s \ln(\zeta^{-1}) \|\eta\|_{X^*} \pl .
\end{align*}
for some constant $c_p$ that depends only on $p$.
\end{theorem}

\begin{proof}
By Lemma~\ref{sym}, we have
\begin{align*}
\ez \si(g)^* \eta \eta^* \si(g)
& = \int_G \si(g)^* \eta \eta^* \si(g) d\mu(g)
= \int_G \si(g^{-1}) \eta \eta^* \si(g) d\mu(g) \\
& = \frac{\mathrm{tr}(\eta\eta^*)}{N} \mathrm{Id}
= \frac{\|\eta\|_2^2}{N} \mathrm{Id} = \mathrm{Id} \pl .
\end{align*}
Furthermore, since $\si(g)(B_X)\subset B_X$ we deduce that
\[
\ez \sup_{1 \leq j \leq m} \sup_{x\in B_X} |\langle \eta,\si(g_j)x \rangle|^{2q}
\leq \|\eta\|_{X^*}^{2q} \pl, \quad \forall q \in \mathbb{N} \pl .
\]
Lastly, Lemma~\ref{eduality} implies that $M_{p,\al_d}(K)\kl (1/p-1/2)^{-1} (CT_p(X^*))^{p'+1}$. Hence the assertion follows from Theorem~\ref{main:rv}.
\qd

\begin{exam}
Schatten class $S_q^n$ of $n$-by-$n$ square matrices has enough symmetries. Therefore, Theorem~\ref{point} provides an alternative proof for a near optimal RIP of partial Pauli measurements by Liu \cite{liu2011universal}. Pauli measurements are given as an orbit of the Clifford group with $\eta = \sqrt{n} \mathrm{Id}_{\cz^n}$. Since $S_\infty^n$ is not type 2, let $X$ be $S_q^n$ with $q' = \ln n$ instead of $S_1^n$. Obviously, all rank-$s$ matrices is $(K,s)$-sparse with $K = B_{S_q^n}$ in our generalized sparsity model. Then $T_2(X^*) \kl \sqrt{\ln n}$. In other words, the complexity of $K$ is a logarithmic term. On the other hand, the incoherence satisfies $\|\eta\|_{S_{q'}} \leq e \|\eta\|_{S_\infty} = e \sqrt{n}$ and the upper bound is proportional to $n$. This large incoherence is the penalty for noncommutativity.
\end{exam}

\section{RIP on low-rank tensors} 
\label{sec:tensor}

In this section, we apply the main results in previous sections to demonstrate that the group structured measurement can be useful for dimensionality reduction of higher-order low-rank tensors.
Let $N,n,d \in \mathbb{N}$ satisfy $N = n^d$. We consider the convex set $K\subset B_2^N$ given by the convex hull of rank-1 tensors, i.e.
\begin{equation}
\label{Ktensor}
K = \absc\{ y_1\ten \cdots \ten y_d \pl|\pl y_j \in B_2^n, \pl \forall j=1,\dots,d \} \pl.
\end{equation}
Note that $K$ is the unit ball of $(\ell_2^n)^{\ten_{\pi}^d}$, which is the tensor product of $d$ copies of $\ell_2^n$ with respect to the largest tensor norm $\pi$. Let $G$ be any compact group with an affine isotropic action, then the product $G^d$ admits an affine isotropic action which leaves $K$ invariant. For example, the tensor product representation $\zz_{n^{2d}}$ of the quantum Fourier transform shows that $K$ has enough symmetries.

\begin{lemma}\label{count}
Let $K$ be defined in \eqref{Ktensor}. There exist rank-1 tensors $x_1,\dots,x_M \in B_2^N$ with $\ln M\kl 3nd(1+\ln d)$ such that $K \subset e \absc\{x_j \pl|\pl 1\le j\le M\}$.
\end{lemma}

\begin{proof}
Let $0<\eps<1/2$ and $\Delta$ be an $\eps$-net for the unit ball $B_2^n$. Then we may assume that $|\Delta|\le (1+2/\eps)^n$ (we consider the real scalar field). It follows that every element $y \in B_2^n$ has a representation
\[ y \lel \sum_{j=1}^{\infty} \al_j z_j \]
with $z_j\in \Delta$ and $\sum_j |\al_j|\kl \frac{1+\eps}{1-\eps}\kl (1+3\eps)$. This implies
\[
y_1\ten \cdots \ten y_d
\lel \sum_{j_1,\dots,j_d=1}^{\infty} \Motimes_{l=1}^d \al_{j_l} z_{j_l} \pl.
\]
Note that $\Motimes_{l=1}^d z_{j_l} \in \Delta^{\ten^d}$ and
\[
\sum_{j_1,\dots,j_d=1}^{\infty} \Big|\prod_{l=1}^d \al_{j_l}\Big|
\kl (1+3\eps)^d \pl .
\]
Thus we choose $1/(d+1)<3\eps\le 1/d$ and deduce the assertion from
\begin{align*}
\ln |\Delta^{\ten^d}|
&\le nd \ln\left(1+\frac{2}{\eps}\right)
\kl nd\ln 13d \pl . \qedhere
\end{align*}
\qd

\begin{cor}\label{epsilon}
Let $X$ be the Banach space with the unit ball $K$ defined in \eqref{Ktensor}, $0<\zeta<1$, and $\bm{\xi} \sim \mathcal{N}(0,I_N)$. Then for all $r \in \mathbb{N}$ we have
\begin{equation}
\label{ubdualnorm_tensor}
(\ez \|\bm{\xi}\|_{X^*}^r)^{1/r} \kl c \max\{ \sqrt{r},\sqrt{3nd(1+\ln d)}\} \pl .
\end{equation}
for a numerical constant $c$. Furthermore,
\[
\mathrm{Prob}\Big(\|\bm{\xi}\|_{X^*} \geq \sqrt{2(1+3nd(1+\ln d)+\ln(\zeta^{-1}))} \Big) \leq \zeta.
\]
Here $\|\cdot\|_{X^*}$ denotes the dual norm of $X$.
\end{cor}

\begin{proof}
By Lemma \ref{count}, there exists $\Delta \subset B_2^N$ such that $K \subset e\absc \Delta$ and $\ln |\Delta| \le 3nd(1+\ln d)$. Then
\[
\|\bm{\xi}\|_{X^*} \lel \sup_{x \in K} |\langle x,\bm{\xi}\rangle| \kl e \sup_{x \in \Delta} |\langle x,\bm{\xi}\rangle| \pl .
\]

Let $p=3nd \ln d$ and $r\gl p$. Since the net $\Delta$ is contained in the unit ball we deduce
\begin{align*}
(\ez \|\bm{\xi}\|_{X^*}^r)^{1/r}
\kl e (\ez \sup_{x\in \Delta} |\langle\bm{\xi},x\rangle|^r)^{1/r}
\kl e |\Delta|^{1/r} \sup_{x\in \Delta} (\ez |(\bm{\xi},x)|^r)^{1/r} \kl e^2 \sqrt{r}  \pl .
\end{align*}
For $r\le p$ we just use the $L_p$ norm.
The second assertion follows by the union bound over $\Delta$.
\qd

\begin{theorem}
\label{tensor}
Let $N=n^d$, $K$ be given in \eqref{Ktensor}, $K_s = \sqrt{s}K \cap \mathbb{S}^{N-1}$, and $\bm{\xi} ~ \mathcal{N}(0,I_N)$. Let $G$ be a compact group with an affine isotropic action, $g_1,\dots,g_m$ be independent random variables in $G^d$ with respect to the Haar measure, and $\si:G^d\to O_N$ be a tensor product representation.
Let $\delta > 0$ and $0<\zeta<1$. Then
\[
\mathbb{P}\left(
\sup_{x\in K_s} \Big|\frac{1}{m}\sum_{j=1}^m |\langle \si(g_j)^*\bm{\xi},x\rangle|^2 - \|x\|_2^2\Big| \geq \max(\delta,\delta^2)
\right) \leq \zeta
\]
provided that
\[
m \gl c \delta^{-2} s (1+\ln m)^3 (1+3nd(1+\ln d)+\ln(\zeta^{-1}))^2 \pl
\]
holds for a numerical constant $c$.
\end{theorem}

\begin{proof}
As in the proof of Lemma~\ref{epsilon}, construct $\Delta \subset B_2^N$ from Lemma~\ref{count} such that $\ln |\Delta| \leq 3dn(1+\ln d)$ and $\widetilde{K} = e \absc \Delta$ contains $K$. Let $\widetilde{X}$ be the Banach space induced from $\widetilde{K}$ such that the unit ball in $\widetilde{X}$ is $\widetilde{K}$. Since $K \subset \widetilde{K}$, it follows that $K_s \subset \sqrt{s} \widetilde{K} \cap \mathbb{S}^{N-1}$. Moreover, we have $\|x\|_{X^*} \leq \|x\|_{\widetilde{X}^*}$ for all $x$, where $\|\cdot\|_{\widetilde{X}^*}$ denotes the dual norm of $\widetilde{X}$.
Therefore by Theorem~\ref{randrv} the assertion holds if
\[
m \gl c \delta^{-2} s \max\Big((1+\ln m)^3 (1+\ln M), \ln(\zeta^{-1}) \Big) \|\eta\|_{\widetilde{X}^*}^2 \pl .
\]
Indeed, according to the proof of Lemma~\ref{epsilon}, the right-hand side of the inequality in \eqref{ubdualnorm_tensor} is also a valid upper bound on $\|\eta\|_{\widetilde{X}^*}$. This completes the proof.
\qd

Let us now compare the estimate in Theorem~\ref{tensor} to the Gaussian measurement operator.
Let $\bm{\xi}_1,\dots,\bm{\xi}_m$ be independent copies of $\bm{\xi} \sim \mathcal{N}(0,I_N)$. Then by Gordon's escape through the mesh \cite[Corollary~1.2]{gordon1988milman} and Lemma~\ref{epsilon}, it follows that
\[
\mathbb{P}\left(
\sup_{x\in K_s} \Big|\frac{1}{m}\sum_{j=1}^m |\langle \bm{\xi}_j,x\rangle|^2 - \|x\|_2^2\Big|
\geq \max(\delta,\delta^2)
\right) \leq \zeta
\]
provided
\[
m \geq c \delta^{-2} \left(nd(1+\ln d)+\ln(\zeta^{-1})\right) \pl.
\]

To simplify the expressions for the number of measurements, let us choose $\zeta$ not too small so that $\ln(\zeta^{-1})$ is dominated by the other logarithmic terms and then ignore the logarithmic terms. While the Gaussian measurement operator provides the RIP with roughly $snd$ measurements, the group structured measurement with a Gaussian instrument provides the RIP roughly with $sn^2d^2$ measurements. However, the suboptimal scaling of $m$ for the group structured measurement can be compensated by applying a Gaussian matrix to the obtained measurements (see \cite[Theorem~4.2]{junge2016ripII}). Since $m \approx sn^2d^2$ is already significantly small compared to the dimension $n^d$ of the ambient space $(\ell_2^n)^{\ten_d^\pi}$, this two step measurement system is much more practical than the measurement system with a single big Gaussian matrix.

Moreover, besides suboptimal scaling of the number of measurements for the RIP, the group structured measurement operator has the following advantages. The transformations $\si(g)$ preserve both the convex body $K$ and the $\|\pl\|_2$ norm. The incoherence in this case of group structured measurement is determined by the instrument. Lemma \ref{epsilon} suggests that a random gaussian vector $\bm{\xi}$ can be a good choice for the instrument in the sense that it makes the incoherence parameter small. There exist fast implementations for certain group action transforms, which enable highly scalable dimensionality reduction for massively sized tensor data.

\section*{Acknowledgement}
This work was supported in part by NSF grants IIS 14-47879 and DMS 15-01103.
The authors thank Yihong Wu and Yoram Bresler for helpful discussions.



\end{document}